\documentclass[journal]{IEEEtran}

\usepackage[linesnumbered,lined,noend,ruled]{algorithm2e}
\usepackage{algorithmic}
\usepackage{amsmath,amssymb}
\usepackage{subfigure}
\usepackage{amstext}
\usepackage{amsfonts}
\usepackage{bbm}
\usepackage{color}
\usepackage{amssymb}
\usepackage{float}

\usepackage{color}
\definecolor{red}{RGB}{255,0,0}

\usepackage{multirow}
\usepackage[export]{adjustbox}
\usepackage{ragged2e}
\usepackage{array}

\usepackage{graphicx}
\usepackage{amsmath}
\usepackage{amssymb}
\usepackage{booktabs}
\usepackage{multirow}
\usepackage{float}
\usepackage{dsfont}

\usepackage{xspace}

\newcommand{\eg}{\textit{e.g.}\xspace}
\newcommand{\etal}{\textit{et al.}\xspace}
\newcommand{\etc}{\textit{etc.}\xspace}

\begin{document}
%
\title{Inverse-like Antagonistic Scene Text Spotting via Reading-Order
Estimation and Dynamic Sampling}
%
%
%

\author{Shi-Xue Zhang, Chun Yang, Xiaobin Zhu, Hongyang Zhou, Hongfa Wang, Xu-Cheng Yin

\thanks{Corresponding authors: Xiaobin Zhu.}
\thanks{Shi-Xue Zhang, Chun Yang, Xiaobin, Zhu, Hongyang Zhou and Xu-Cheng Yin are with the School of Computer and Communication Engineering, University of Science and Technology Beijing (USTB)}
\thanks{Hongfa Wang  is with Tencent Technology (Shenzhen) Co. Ltd; (e-mail: hongfawang@tencent.com).}
\thanks{Manuscript received *** **, 2023; revised *** **, 2023.}}

%
%

\markboth{Journal of \LaTeX\ Class Files,~Vol.~14, No.~8, August~2021}%
{Shell \MakeLowercase{\textit{et al.}}: Bare Demo of IEEEtran.cls for IEEE Journals}
%



\maketitle

\begin{abstract}
Scene text spotting is a challenging task, especially for inverse-like scene text, which has complex layouts, \eg, mirrored, symmetrical, or retro-flexed. In this paper, we propose a unified end-to-end trainable inverse-like antagonistic text spotting framework dubbed IATS,  which can effectively spot inverse-like scene texts without sacrificing general ones. Specifically, we propose an innovative reading-order estimation module (REM) that extracts reading-order information from the initial text boundary generated by an initial boundary module (IBM). To optimize and train REM, we propose a joint reading-order estimation loss ($ \mathcal{L}_{RE} $) consisting of a classification loss, an orthogonality loss, and a distribution loss. With the help of IBM, we can divide the initial text boundary into two symmetric control points and iteratively refine the new text boundary using a lightweight boundary refinement module (BRM) for adapting to various shapes and scales. To alleviate the incompatibility between text detection and recognition, we propose a dynamic sampling module (DSM) with a thin-plate spline that can dynamically sample appropriate features for recognition in the detected text region. Without extra supervision, the DSM can proactively learn to sample appropriate features for text recognition through the gradient returned by the recognition module. Extensive experiments on both challenging scene text and inverse-like scene text datasets demonstrate that our method achieves superior performance both on irregular and inverse-like text spotting.
\end{abstract}


\IEEEpeerreviewmaketitle

\section{Introduction}\label{sec:introduction}
\begin{figure}[tp]
	\begin{minipage}[t]{0.99\linewidth}
		\includegraphics[width=1\linewidth]{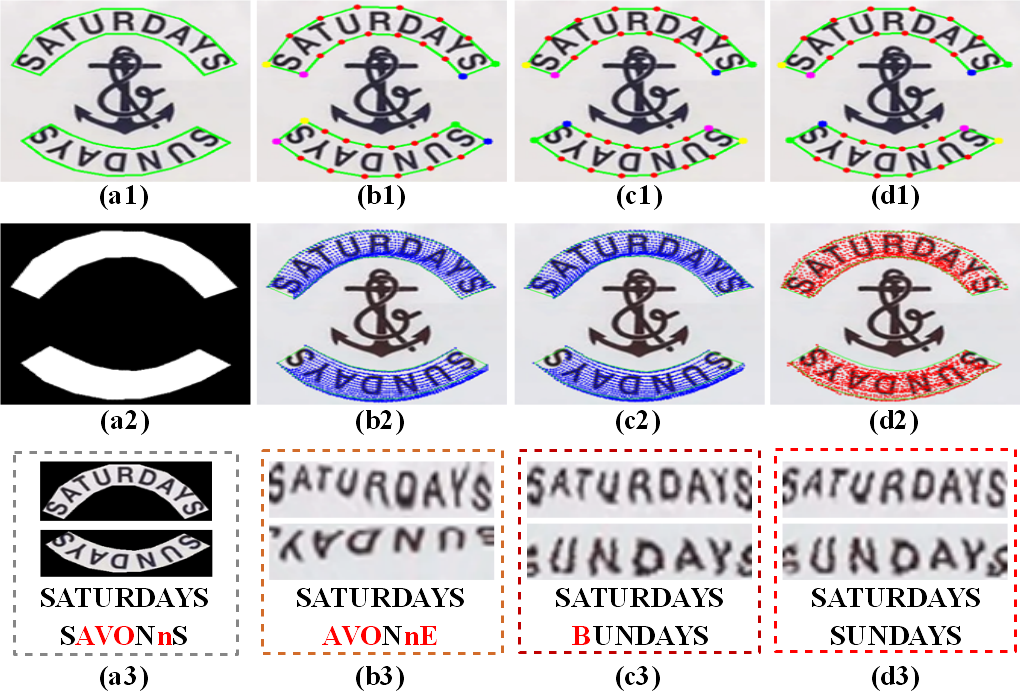}
		\caption{Illustrations of different text feature sampling methods: (a) Masked RoI: a shape mask is used to formulate text regions, while background noise can be suppressed; (b)  TPS without reading-order: the text is transformed into a horizontal region using boundary control points to generate fixed sample grids; (c) TPS with reading-order; (d) DSM with reading-order.} \label{fig:intro1}
	\end{minipage}%
	\vspace{-1em}
\end{figure}

\IEEEPARstart{T}{ext} spotting aims to localize and recognize text in images. It has received ever-increasing attention for its extensive real-world applications, such as vehicle intelligence and road sign recognition in autonomous driving. Although text spotting has made significant progress recently, existing methods still face challenges in recognizing text with complex layouts, such as arbitrary orientations or shapes~\cite{PAN++, TextDragon, Boundary, MANGO,GLASS}. To address these problems, existing text spotting frameworks propose the Masked RoI~\cite{PAN++, Mask_TextSpotter_v3,TUTS} and Thin Plate Spline (TPS)~\cite{TextPerception, Boundary, TPSNet} strategies. Masked RoI based methods allow for the suppression of background information, as shown in Fig.~\ref{fig:intro1} (a1-a2), but they still struggles with irregular texts. TPS based methods can transform irregular texts into horizontal texts by symmetrical boundary points, as seen in Fig.~\ref{fig:intro1} (b1-b3), but they also suffers from the accuracy of detected boundaries. 

In text spotting, two critical problems still require further improvement. Firstly, the crucial reading-order information is not fully explored for the recognizer to decode text characters in the correct sequence. Although many existing datasets follow the text reading direction, most ignore this issue, except Text Perceptron~\cite{TextPerception} and PGNet~\cite{PGNet}. However, these two approaches only use information pointing to the head and tail of text, which in some cases (as shown in Fig.~\ref{fig:intro1}(a1)) cannot fully reflect the reading-order. Inverse-like texts are universal and appear mirrored, symmetrical, or retroflexed. Simply clipping the detection result of such exceptional text makes recognition difficult, as shown in Fig.~\ref{fig:intro1} (a1-a3). Suppose the network can fully excavate and learn the reading-order information hidden in the training samples labeled with the reading direction. In that case, the reading-order information will suitably align the text for better recognition.
The second issue is that text recognition accuracy is heavily dependent on the precision of detection, resulting in potential error propagation from detection to recognition, as pointed out in previous studies \cite{DEER, SRSTS}. 
Current methods ~\cite{PAN++, TextPerception} generally adopt inflexible sampling strategies that rely on fixed sampling grids manually determined by detection boundaries or segmented masks. As a result, when detection accuracy is compromised or sampling features are inadequate, the recognizer may fail in decoding the correct sequence, as shown in Fig.~\ref{fig:intro1} (c1-c3). Therefore, exploring an adaptive dynamic feature sampling approach that can improve recognition performance is nontrivial, especially in challenging scenarios where detection accuracy is limited.

\begin{figure}[tp]
	\begin{minipage}[t]{0.99\linewidth}
		\includegraphics[width=1\linewidth]{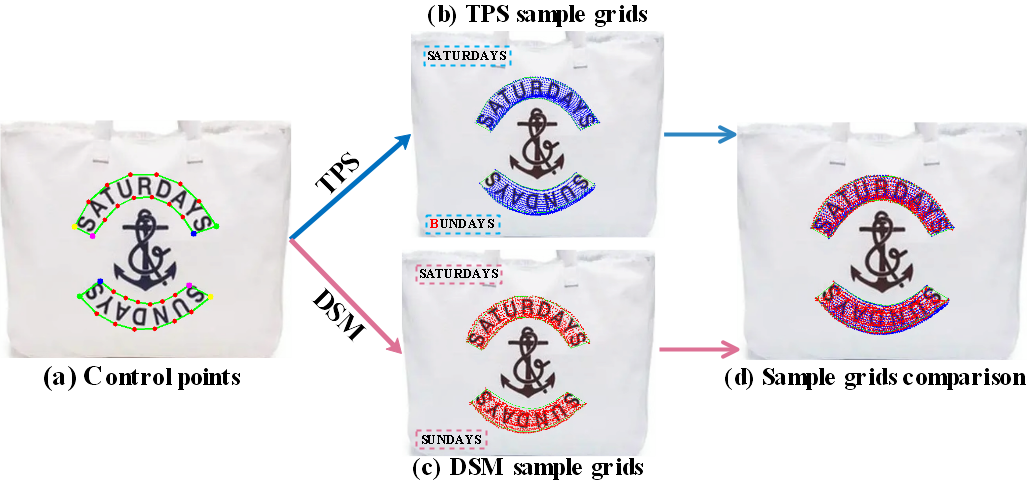}
		\caption{Comparison of TPS and DSM.(a) Boundary control points with reading-order; (b) TPS: generating a fixed and regularly sampling grids heavily relay on control points; (c) DSM:  dynamically generating adaptive sampling grids through self-adjustment with recognition model; (d) Visual comparison of sample grids for TPS and DSM.} \label{fig:intro2}
	\end{minipage}%
	\vspace{-1.5em}
\end{figure}

In this paper, we propose a unified end-to-end trainable inverse-like antagonistic text spotting framework (dubbed \textbf{IATS}), which can effectively spot inverse-like scene texts without sacrificing general ones following human reading habits. Specifically, we propose an innovative reading-order estimation module (REM) that extracts reading-order information from the initial text boundary generated by an initial boundary module (IBM). As shown in Fig.~\ref{fig:intro2} (a), the reading-order estimation module (REM) based on a circular convolution network can accurately estimate four key corner points on the coarse initial boundary. To ensure the reliability of REM, we propose a novel joint reading-order estimation loss ($ \mathcal{L}_{RE} $) to optimize and train REM, which includes a classification loss, an orthogonality loss, and a distribution loss. According to the predicted reading-order, we divide the initial text boundary into two symmetric boundary control points and use a lightweight boundary refinement module (BRM) to iteratively refine them to adapt to the diversity of text shapes and scales. To further alleviate the incompatibility between detection and recognition, we propose a novel dynamic sampling module (DSM) with thin-plate spline, which is used to dynamically sample features in the detected text region for recognition. During training, DSM can actively learn how to dynamically sample optimal features through the gradient returned by the recognition module without extra supervision. Benefiting from the DSM, our method can recognize text instances accurately even if detected text boundaries are not perfect, as shown in Fig.~\ref{fig:intro1} (d1-d3). Extensive experiments on challenging scene text (Total-Text, CTW-1500, and ICDAR2015) and inverse-like scene text datasets (Rot.Total-Text, and Inverse-Text) verify that our method achieves superior performance both on irregular and inverse-like text spotting tasks.

Overall, our main contributions are summarized as follows:

\begin{itemize}
   \item We propose a unified end-to-end trainable inverse-like antagonistic text spotting framework (dubbed \textbf{IAST}),  which can effectively spot inverse-like scene texts without sacrificing general ones. 
	
   \item We propose an innovative reading-order estimation module (REM) with a joint reading-order estimation loss ($ \mathcal{L}_{RE} $) to fully learn and excavate the key reading-order information in text boundaries.
	
  \item We propose a dynamic sampling module (DSM), which can adaptively learn how to dynamically sample appropriate features in detected text regions through the gradient returned by the recognition module. 
	
  \item Extensive experiments verify that our method achieves competitive results in scene text spotting benchmarks and also significantly surpasses previous methods in spotting irregular and inverse-like scene text.
	
\end{itemize}

\begin{figure*}[htbp]
	\begin{center}
	\includegraphics[width=1.0\linewidth]{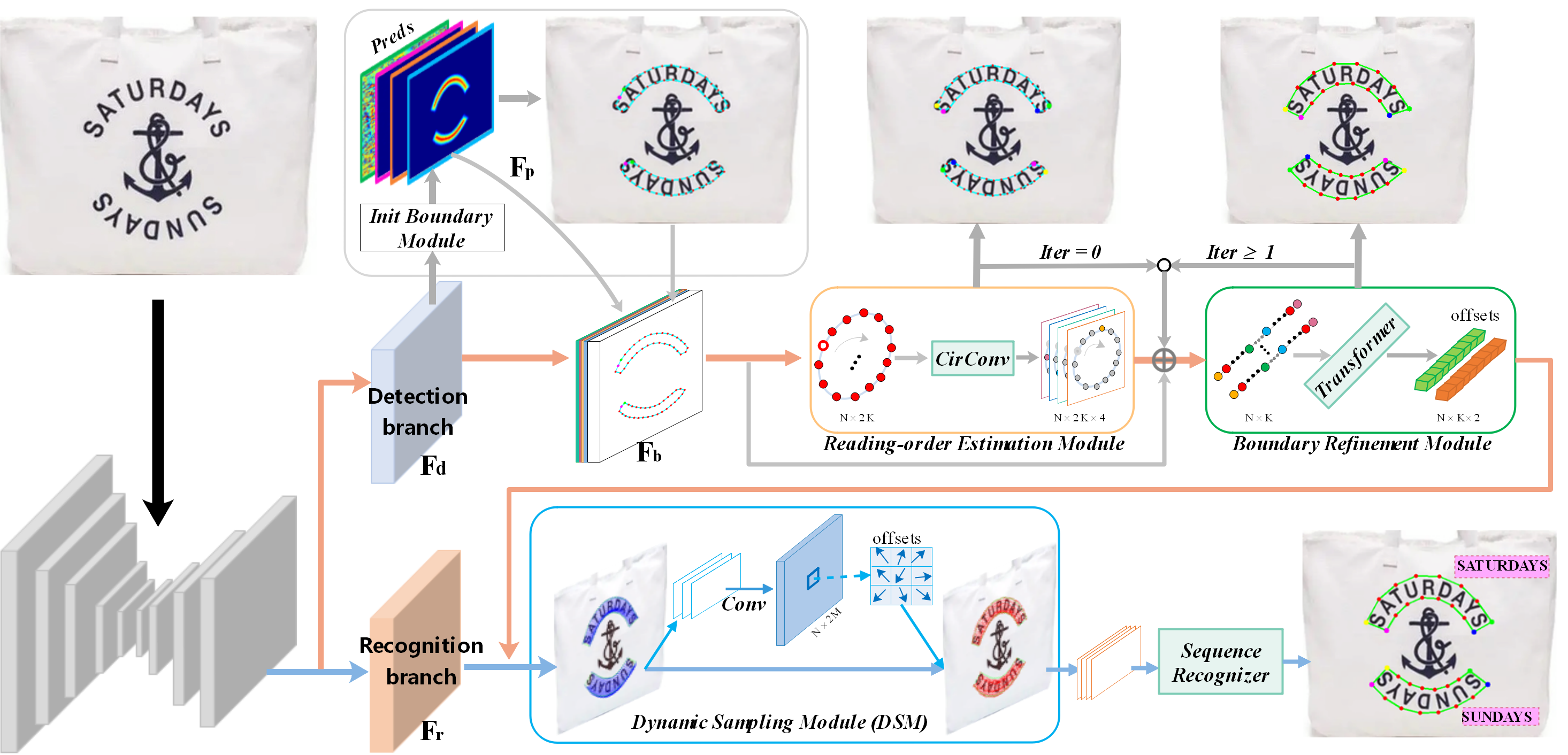}
	\caption{Overview of the proposed framework. The orange lines indicate the detection flow, and the blue lines indicate the recognition flow. The predictions of the different modules are also visualized in the origin image.}
	\label{fig:framework}
	\end{center}
\vspace{-1.8em}
\end{figure*}

\section{Related Work} \label{Related_Work}

\subsection{Text Detection} 
Traditional deep learning-based text detection methods~\cite{RRPN,HAM, Zhang2022GraphFN, MOST, TextPMs, HGR-Net, CM-Net} mainly focus on multi-oriented texts. Anchor-based methods~\cite{RRPN} adopt rotated anchors and RRoI pooling for detecting multi-oriented texts. Anchor-free methods~\cite{MOST} directly regress the offsets from boundaries or vertexes to the current point for detecting texts. Some methods~\cite{HAM, AAM} try to design a hidden anchor mechanism to integrate the advantages of the anchor-based method into the anchor-free method. Recently, a series of text detection methods have been proposed for detecting irregular text. The Connected Component (CC) based methods~\cite{DRRG,SegLink++} usually detect individual text parts or characters first, followed by a link or group post-processing procedure for generating final texts. Segmentation-based methods~\cite{KPN, TextPMs}  
use instance segmentation to detect arbitrary shape text and design different schemes to separate adjacent text instances. But, segmentation accuracy significantly determines the quality of detected boundaries.
Contour-based methods~\cite{FCENet,TextBPN, TextRay, DPText, TextBPN++} resort to modeling the text boundary for better representation of arbitrarily-shaped
texts.
 
\vspace{-0.5em}
\subsection{Text Recognition}
Scene text recognition involves recognizing texts in a cropped image. Traditional methods~\cite{PhotoOCR,E2E} rely on character-level annotations for character detection, while methods~\cite{CRNN1, CRNN2} extract features from line-level text using CNN and RNN, and use a CTC-based decoder for prediction alignment. However, these methods are designed for regular text recognition and struggle with irregular text. To address this problem, Shi \etal~\cite{STN1} propose a rectification network with STN for arbitrary shape text, while Litman \etal~\cite{SCATTER} use TPS and selective attention decoder for visual and contextual features. CharNet~\cite{CharNet} uses a Character-Aware Neural Network to detect characters first and then separately transform them into a horizontal one. AON~\cite{AON} extracts features with four directions and character position clues, while SAD~\cite{SAD} applies a 2D-attention mechanism to catch irregular text features, both achieving impressive results. To address the attention drift issue, RobustScanner~\cite{RobustScanner} design a position enhancement branch in the recognition model. In addition, some methods use semantic segmentation to assist in text recognition.

\vspace{-0.5em}
\subsection{Text Spotting}
Traditional text spotting methods~\cite{textboxes, textboxes++} perform text detection and recognition as two separate steps. Generally, a text detector extracts regions of interest (RoI), which are then fed into a recognition model. However, recent end-to-end text spotting approaches~\cite{FOTS, MaskTextSpotter, ABCNet, TextPerception} have confirmed that detection and recognition tasks are highly relevant and complementary to each other. By sharing features and jointly optimizing the modules \cite{FOTS, Deep_TextSpotter} in a unified end-to-end trainable network, they achieved improved detection and recognition performances simultaneously.

Recently, several methods~\cite{TextDragon,Boundary_TextSpotter,ABCNet, Mask_TextSpotter_v3, PAN++, CRAFTs,PGNet,DPText,TESTR, SPTS, DeepSolo, SPTS_v2} have been proposed to address arbitrary shape text spotting. Mask TextSpotter~\cite{Mask_TextSpotter_v3} and PAN++ \cite{PAN++} use RoI Masking to focus on the arbitrarily shaped text region. MANGO \cite{MANGO} uses a Mask Attention module to retain global features for multiple instances but still requires centerline segmentation to guide the grouping of the predictions. Boundary TextSpotter~\cite{Boundary_TextSpotter} and TPSNet~\cite{TPSNet} localize the boundary points of text instances and rectify the features using Thin-Plate-Spline before feeding them into the recognition branch. CRAFTS~\cite{CRAFTs} uses character region maps supervised by character-level annotations to help the attention-based recognizer attend to precise character center points. PGNet~\cite{PGNet} transforms the polygonal text boundaries to the centerline, border offset, and direction offset and performs multi-task learning for these objectives. Inspired by Pix2Seq~\cite{Pix2seq}, some methods, such as TESTR~\cite{TESTR}, TTS~\cite{TTS}, and SPTS~\cite{SPTS}, use a network combining CNN and Transformer that tackles text spotting as a sequence prediction task, similar to language modeling. However, these methods usually require extensive computing and data resources.

Although the text spotting methods (\eg, ABCNet~\cite{ABCNet,ABCNet_v2}, TESTR~\cite{TESTR} and SwinTextSpotter~\cite{Swin-Transformer}) have achieved great improvement for arbitrarily shaped text spotting, they still suffer from inverse-like text because of the absence of key reading-order information. However, the reading of inverse-like scene text is important, even though it has just been noticed in DPText~\cite{DPText}. The other problem is that the reading performance of these methods also suffers from the accuracy of detected boundaries because of fixed feature sample. In this paper, we aim to effectively spot inverse-like scene texts without sacrificing general ones. Hence, we design a reading-order estimation module and a dynamic sampling module, which greatly improves the accuracy of inverse-like text spotting without losing generality.


%
%

\section{Proposed Method} \label{Proposed_Method}
\subsection{Overview}
The framework of our method presented in Fig.~\ref{fig:framework} mainly consists of six components: feature extraction module, init boundary module (IBM), reading-order estimation module (REM), boundary deformation module, dynamic sampling module (DSM), and text recognition module. The feature extraction module extracts features from input images for the text detection and recognition tasks. To preserve spatial resolution and utilize multi-level information, we use a multi-level feature fusion strategy similar to~\cite{FPN}. As noted in~\cite{Boundary_TextSpotter}, the two tasks have different requirements for feature maps. Specifically, the recognition task needs more detailed information for character sequence prediction, while the detection task focuses on the whole text instance. Therefore, we use lightweight convolutions to separate the detection and recognition features.

\begin{figure}[tbp]
	\begin{center}
	\includegraphics[width=0.96\linewidth]{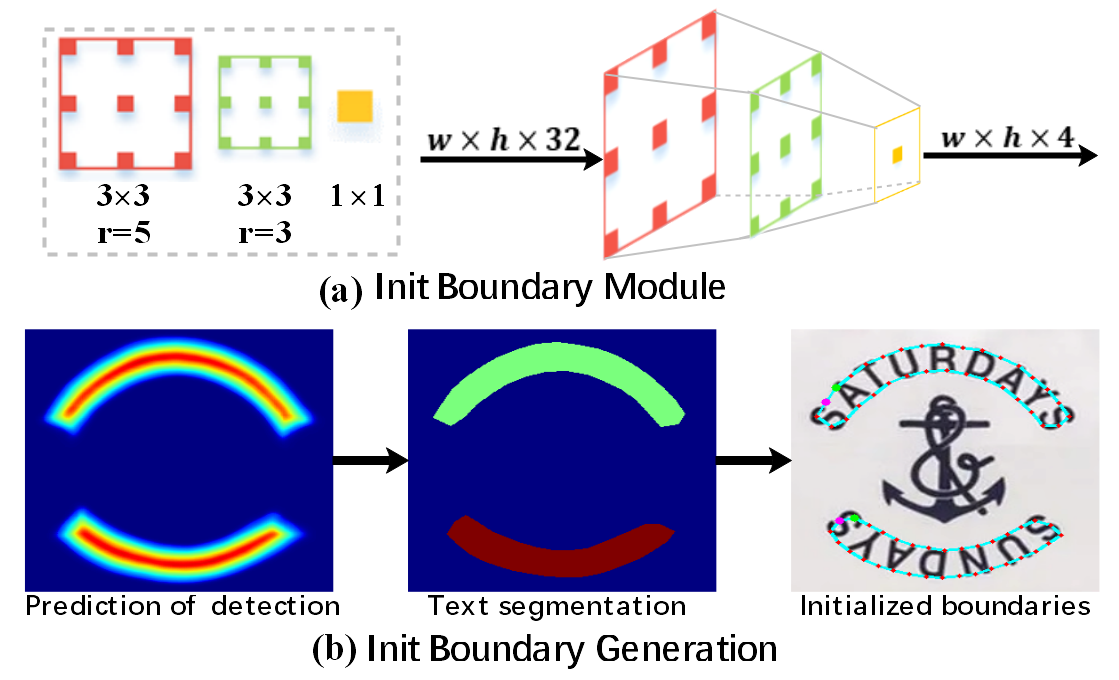}
		\caption{(a) Architecture of initial boundary module; (b) The generation of initial text boundaries.}	
	\label{fig:ibm}
	\end{center}%
\vspace{-1.8em}
\end{figure}

\subsection{Initial Text Boundary}
The initial boundary module generates rough text boundaries to locate text instances. Similar to~\cite{TextBPN, TextBPN++}, our module consists of multi-layer dilated convolutions, including two $3 \times 3$ convolution layers with different dilation rates and one $3 \times 3$ convolution layer, as illustrated in Fig.~\ref{fig:ibm}(a). It uses shared detection features to provide prior information for text location. The prior information can be in the form of field information such as classification, distance, and direction in\cite{TextBPN, TextBPN++}, probability maps in DB~\cite{DB}, or text kernels in PSENet~\cite{CVPR19_PSENet} and PAN~\cite{PSENet_v2}. Using the prior information, we can obtain the text segmentation and generate a coarse text boundary, as shown in Fig.~\ref{fig:ibm}(b). The accuracy of these text boundaries is not deterministic, as they are only used as initial information and will be refined in our method.

\subsection{Reading-Order Estimation}
\begin{figure}[tbp]
	\begin{center}
	\includegraphics[width=0.96\linewidth]{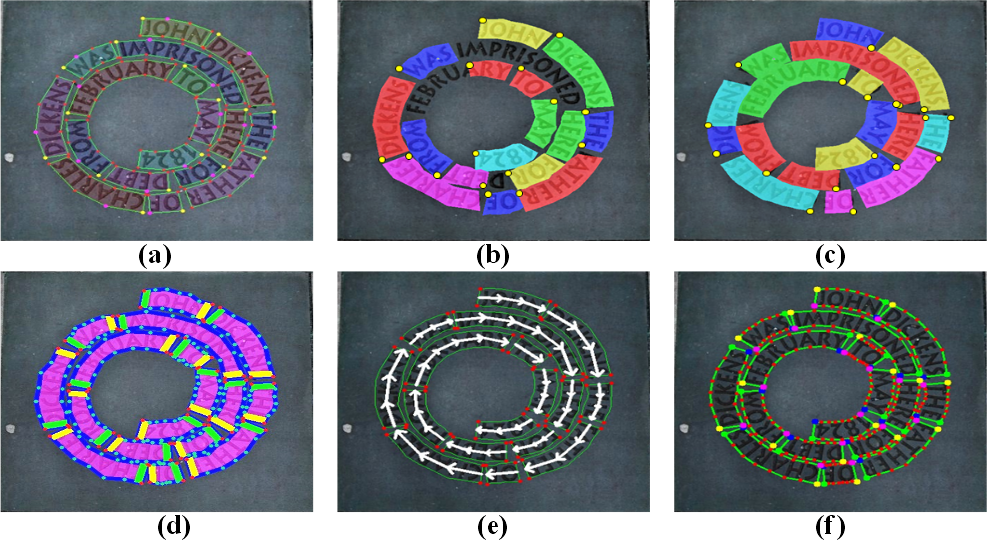}
		\caption{(a) The original label form of text boundary implies the reading-order. (b) The original label form induces the detector to implicitly learn the reading order, resulting in false positives and jagged edges. (c) Even with extensive rotation augmentation during training, the detector still can't learn the reading-order well. (d) Text Perceptron~\cite{TextPerception} uses order-aware segmentation to indicate the head and tail of text instances and capture latent reading-orders. (e) PGNet~\cite{PGNet} uses text direction offset (TDO) maps to extract the text reading-order. (f) Our method uses the four key corners on the text boundary to indicate the reading-order.}	
	\label{fig:diff_order}
	\end{center}%
\vspace{-1.8em}
\end{figure}

The reading-order is crucial for accurate understanding and recognition of text sequences. In existing text detection datasets, the annotation of text detection follows the reading direction of human beings, as it aligns the text features suitably for better recognition. However, the implicit learning of reading-order can degrade the detector's robustness, resulting in false positives and jagged edges, as shown in Fig.~\ref{fig:diff_order} (b). Even with extensive training with rotation augmentations, the detector still struggles to learn correct reading-orders, as shown in Fig.~\ref{fig:diff_order} (c).

Learning reading-order explicitly is rare in existing methods, except for Text Perceptron~\cite{TextPerception} and PGNet~\cite{PGNet}. Text Perceptron uses order-aware segmentation to indicate the head and tail of text instance, while PGNet extracts the text reading-order using TDO maps, as shown in Fig.~\ref{fig:diff_order} (d) and (e). However, both methods only consider the order of information from the head to tail, neglecting the top and bottom. In some cases, this imperfect reading-order fails to reflect the actual case, such as inverse-like text, which may occur in natural images with more complex layouts like mirror, symmetry, or inversion, \etc. In these cases, simply clipping the detection results with an artificial or imperfect reading-order disturbs recognition, as shown in Fig.~\ref{fig:intro1} (a1-a3) or Fig.~\ref{fig:diff_order} (b-e).

\begin{figure}[htbp]
	\begin{center}
		\includegraphics[width=0.99\linewidth]{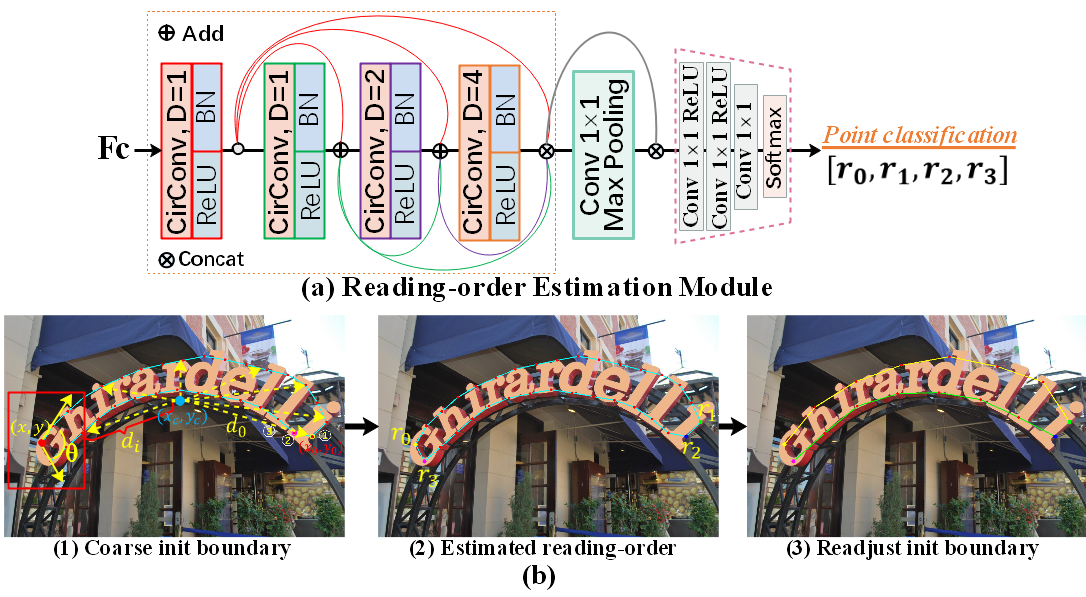}
		\caption{(a) The structure of Reading-order Estimation Module (REM). (b) The adjustment process of the initial boundary with predicted reading-order.}	
		\label{fig:order_model}
	\end{center}%
	\vspace{-1.8em}
\end{figure}

\begin{figure*}[tbp]
	\begin{center}		
        \includegraphics[width=0.95\linewidth]{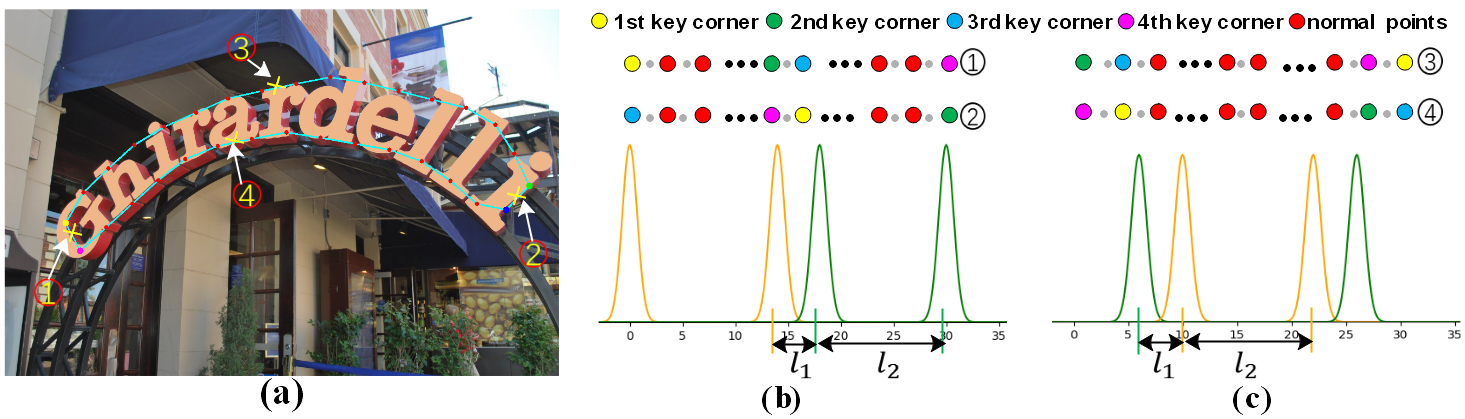}
		\caption{(a) The $ 2K $ control points on detected text boundary, where red points represents normal control points, and non-red points (yellow, green, blue, and purple) represent four key corners for reading-order. (b)-(c) The distribution of key corners in these $ 2K $ control points can be divided into two different types when these control points are arranged into a one-dimensional sequence. The wave with same color indicate the key corner located on the same long side, such as yellow and green, blue and purple.}	
		\label{fig:order_KL}
	\end{center}%
	\vspace{-1.8em}
\end{figure*}

To fully explore the reading-order information of humans hidden in annotations \cite{DPText}, we design a reading-order estimation module (REM) to accurately learn the text reading-order information. In detection tasks, text instances are typically found in long strips, with two long edges in the text boundary, as noted in~\cite{TextSnake,DPText}. Hence, we adopt four key corners at both ends of the text's long side to indicate the reading-order, as depicted in Fig.~\ref{fig:diff_order} (f). In this work, we tackle the problem of reading-order estimation by treating it as a classification task. Our approach involves a reading-order estimation module (REM) that identifies whether each point on the initial boundary belongs to one of the four key corners. Specifically, we employ a ``CirConv" block, a fusion block, and a prediction head (as depicted in Fig.~\ref{fig:order_model} (a)) to construct REM. The ``CirConv" block consists of four circular convolution layers with different dilation rates (\eg, [1, 1, 2, 4]) to enhance REM's information aggregation capability. We use dense shortcut connections across all layers to improve the interaction between each layer \cite{DenseNet}. The fusion block uses a 1$\times$1 convolution layer and max pooling to merge information in the ``CirConv" block. Then, the deeply fused features are distributed to each initial boundary point by concatenating their features. Finally, three 1$\times$1 convolution layers with ReLU activation and sigmoid generate the classification information ($ [r_0, r_1, r_2, r_3] $) for each initial boundary point.

To enable batch processing and avoid missing key corners due to a large sampling interval, we sample $2K$ vertices ($P = {p_{_0},...,p{_i},...p{_{_{2K-1}}}}$) for each initial text boundary, forming a closed contour (as shown in Fig.~\ref{fig:order_model} (b)-(1)). Circular convolution will encode the cyclicity of points along the closed contour effectively, building on the success of DeepSnake \cite{DeepSnake}. However, an excellent network structure is not sufficient to learn the correct reading-order, as the input information also plays a decisive role. To ensure REM obtains more reliable information to accurately estimate the reading-order, the input feature $ F_c(i) $ consists of geometric attributes ($ F_g(i) $) of vertex $p_i$, visual features ($ F_b(i) $), and order embedding features ($ F_e(i) $) of the initial boundary. 

The geometric attributes ($ F_g(i) $) of vertex $ p_i $
includes cosine ($ \cos ({\theta}_i) $) of angle, coordinates $ (\Delta x_i, \Delta y_i)$ relative to the initial boundary centroid $ (x_c, y_c)$, and distance ($ d_i $) relative to the $ (x_c, y_c)$. Thus, the geometric attributes ($ F_g(i) $) of vertex $ p_i $ can be formulated as
\begin{gather}
    F_g(i) = [\cos ({\theta}_i), d_i, \Delta x_i, \Delta y_i],
\end{gather}
The $ cos {\theta}_i $ can be calculated as
\begin{gather}
	cos {\theta}_i=  \dfrac{\overrightarrow{p_{_i}p_{_{i+1}}} \cdot \overrightarrow{p_{_i}p_{_{i-1}}} }{\left\vert\overrightarrow{p_{_i}p_{_{i+1}}}\right\vert \left\vert\overrightarrow{p_{_i}p_{_{i-1}}}\right\vert},
\end{gather}
where the angle ($ \theta $) of vertex $ p_i $ is defined as the angle between vector $ \overrightarrow{p_{_i}p_{_{i+1}}} $ and $ \overrightarrow{p_{_i}p_{_{i-1}}} $, as shown in Fig.~\ref{fig:order_model} (b)-(1). The $ (x_c, y_c)$ is the centroid of the initial boundary, formulated as 
\begin{gather}
(x_c, y_c) = (\dfrac{1}{2K}\sum_{i=0}^{2K-1}x_i, \dfrac{1}{2K}\sum_{i=0}^{2K-1}y_i),
\end{gather}
After we get the $ (x_c, y_c)$, the coordinates $ (\Delta x_i, \Delta y_i)$ and the distance ($ d_i $) of vertex $ p_i $ can be calculated as follows:
\begin{gather}
	(\Delta x_i, \Delta y_i) = (\dfrac{x_i-x_c}{w}, \dfrac{y_i-y_c}{h}),\\
	d_i = \dfrac {\sqrt{\left|x_i-x_c \right|^{2} + \left|y_i-y_c \right|^{2}}}{Max(d_j | j \in [0, 2K))},
\end{gather}
where $ w $ and $ h $ are the width and height of the bounding rectangle of the initial boundary ($ P= \{p_{_0},...,p_{_i},...p_{_{2K-1}}\}$). Here, we have obtained three geometric attributes for vertex $p_i$, providing important prior information for vertex classification. 

We know that text instances in word-level and line-level typically have two long sides and two short sides in their boundaries, and the key points we need to identify are usually located at the intersection of the long and short sides. These points have distinct geometric attributes $F_g(i)=[{\theta}{i}, d_i, \Delta x_i, \Delta y_i]$. As shown in Fig.~\ref{fig:order_model} (b)-(1), the angle ${\theta}{i}$ of these key points usually exhibits sudden changes and is significantly smaller than that of other non-key points. The distances $d_i$ from these key points to the centroid $(x_c, y_c)$ are typically greater than those of other points, although not always the smallest. To improve the fusion of visual ($F_b(i)$) and geometric ($F_g(i)$) features during module reasoning, we embed the geometric attributes of each vertex $p_i$ into a high-dimensional space as \cite{DRRG}. Specifically, we apply sine and cosine functions with varying wavelengths to $F_g$. The geometric attribute embedding is calculated as follows:
\begin{gather}
	{GE}_{({F_g}(i), 2j)}=\cos(\dfrac{2\pi \cdot F_g(i)  }{1000^{{2j}/{C}}}), j \in  (0,C/2 - 1) \label{embed1}, \\ 
	{GE}_{(F_g(i), 2j+1)}=\sin(\dfrac{2\pi \cdot F_g(i) }{1000^{{2j}/{C}}}), j \in  (0,C/2 - 1), \label{embed2}
\end{gather}
where $ C $ is the dimension of the embedding vector (emprirically set to 36). The geometric attribute $ F_g(i) $ of vertex $ p_i $ is embedded into a vector $ {GE}_{(F_g(i))} $ of dimension $ C $. Because $ F_g(i) = [{\theta}_{i}, d_i, \Delta x_i, \Delta y_i] $ has four attribute values, the dimension of each attribute scalar is $ C/4 $.

The $ F_b(x_i, y_i) $ includes a 32-D detection shared features $ F_d $ obtained by CNN backbone and 4-D prior features $ F_p $, as 
\begin{gather}
	F_b(x_i, y_i)= F_d(x_i, y_i) \oplus Fp(x_i, y_i),\\
	F_b(i) = F_b(x_i, y_i) - F_b(x_c, y_c),
\end{gather}
where ``$ \oplus $'' denotes concatenation operation. $ F_b(i) $ has 36 dimensions.  $ F_b(x_c, y_c) $ is the visual features of the centroid $ (x_c, y_c) $ in $ F_b $.

As texts vary in scale, the sampled points may be too close to each other, resulting in very little difference between the visual and geometric features. This can cause confusion in the network output. To avoid this, we embed order information ($F_e(i) = Embedding(i,C)$) of the point sequence into the input feature $F_c(i)$ for vertex $p_i$ using an embedding operation in Pytorch. Therefore, the input feature $F_c$ of the reading-order estimation module is calculated as follows:
\begin{gather}
	F_c= \{  {GE}_{(F_g(i))} \oplus F_b(i) \oplus F_e(i)\} _{i=0}^{2K-1} ,
\end{gather}
where $ i  $  denotes the $ i $-th sampling point in initial boundary. The classification results ($ O =[o_0,o_1,o_2, o_3]
^{T} $) of these sampling points will be obtained after the input features are encoded and reasoned by REM. $ o_j = [\hat{p}_{_{0}},...,\hat{p}_{_{i}}, ..., \hat{p}_{_{2K-1}}] $ is a probability distribution, and each element $ \hat{p_{_{i}}} $ represents the probability that the vertex $ p_{_{i}} $ belongs to $ j $-th class of the key points.

\textbf{Joint reading-order estimation loss.} 
To optimize and train our REM, we propose a joint reading-order estimation loss ($ \mathcal{L}_{RE} $), consisting of a classification loss, an orthogonality loss, and a distribution loss. We use the balance
binary cross entropy loss to supervise classification as
\begin{gather}
	\mathcal{L}_{c} = -\frac{1}{2K}\sum_{j=0}^{3} \sum_{i=0}^{2K-1} {y_{_{i}}}(j)\log(\hat{p}_{_{i}}(j)),
 \end{gather}
where $ y  $ represents the predicted labels for the classification of  every control point. Thus, $ y_{_{i}} $ represents the label for $ i $-th control point $ p_{_{i}}$, which is a one-dimensional vector with four  elements in total (as $ y_{_{i}}  = [y_{_{i}}(0),...,y_{_{i}}(j), ..., y_{_{i}}(3)] $). When the control point $ p_{_{i}} $ belongs to $ j $-th key corner point, $ y_{_{i}}(j) =1 $; otherwise $  y_{_{i}}(j) =0 $. Thus, $ \hat{p}_{_{i}}(j) $ is probability of control point $ p_{_{i}} $ belongs to the $ j $-th key corner point.

To ensure that the four key points found are independent and different from each other, we further designed an orthogonality loss and a distribution loss to constrain the optimization of REM. The orthogonality loss can be expressed as
\begin{gather}
	\mathcal{L}_{o} = D_{KL}(O^{T}O, I),
\end{gather}
where $ D_{KL} $ is the KL Divergence loss; $ O^{T}O $ is the similarity matrix of prediction of REM ( $ O =[o_0,o_1,o_2, o_3]^{T} $); $  I $ is an identity matrix of dimension $ 4 \times 4 $. 

As shown in Fig.~\ref{fig:order_KL} (a), the four key points not only have independent positions but also follow a specific spatial distribution. For word-level and line-level text detection, these key points are usually located at the intersection of the long and short sides in long strips of text. When these sampling points are flattened into a sequence from the short or long side (as \textcircled{1} \textcircled{2} \textcircled{3} \textcircled{4}shown in Fig.~\ref{fig:order_KL} (a)), the distributions of the four key points are similar regardless of category, as shown in Fig.~\ref{fig:order_KL} (b) and (c). Typically, the sequence length of the long side ($ l_{2} $) is greater than that of the short side ($ l_{1} $). To fully utilize this distribution information and better constrain the REM to learn and extract the hidden reading order information, we introduce a distribution constraint loss via KL-divergence as
 \begin{gather}
	\mathcal{L}_{d} = D_{KL}(\sum_{j=0}^{3}o_j, \sum_{j=0}^{3}y_j),
\end{gather}
where $ o_j = [\hat{p}_{_{0}},...,\hat{p}_{_{i}}, ..., \hat{p}_{_{2K-1}}] $ is a probability distribution predicted by REM.  $ y_{j} = [y_{_{0}},...,y_{_{i}}, ..., y_{_{2K-1}}] $ is the one hot representation of sample points ($\{p_{_0},...,p_{_i},...p_{_{2K-1}}\}$) label belonging to $ j $-th category. Finally, the proposed
reading-order estimation loss ($ \mathcal{L}_{RE} $) is a combination of  classification loss $ \mathcal{L}_{c} $ , orthogonality loss $ \mathcal{L}_{o} $, and distribution loss $ \mathcal{L}_{d} $, as follows
\begin{gather}
	\mathcal{L}_{RE} = \mathcal{L}_{c} + \alpha*(\mathcal{L}_{o} + \mathcal{L}_{d}), \label{RE}
\end{gather}
where $ \alpha $ is set to 0.1 because orthogonality loss $ \mathcal{L}_{o} $, and distribution loss $ \mathcal{L}_{d} $ only serves as auxiliary constraints.

According to the four key points obtained, we divide the text boundary into top and bottom sides, as shown in Fig.~\ref{fig:order_model} (b). At the same time, we re-sample $ K/2 $ control points according to the principle of equidistance on the top and bottom sides, respectively. As shown in Fig.~\ref{fig:order_model} (b),
the re-adjusted initial text boundary has  $ K $ control points ($ T = \{c_{_0},..., c_{_i},..., c_{_{K-1}}\} $), which are symmetrically distributed on top and bottom sides.

\begin{figure}[tbp]
	\begin{center}
		\includegraphics[width=0.99\linewidth]{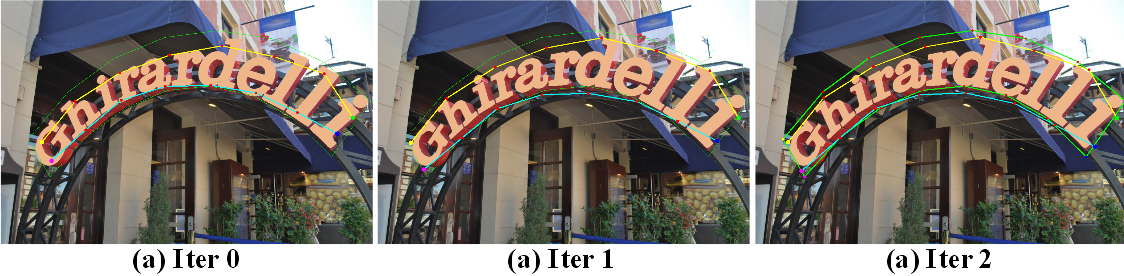}
		\caption{(a)The readjusted initial boundary in $ 0 $-th iteration. (b) The refined text boundary in $ 1 $-th iteration. (c) The refined text boundary in $ 2 $-th iteration. The green contour is the annotated text boundary. The yellow contour is the top side, and the blue contour is the bottom side.}	
		\label{fig:brm}
	\end{center}%
	\vspace{-1.5em}
\end{figure}

\textbf{Boundary Refinement Module.} Due to the varying directions and shapes of text, it is challenging to directly predict accurate text boundary. Therefore, we first use the initial boundary to roughly locate text and separate the neighboring text. But, these coarse initial boundaries  only partially cover the text instances, resulting in inaccurate recognition due to incomplete text regions.  Similar to TextBPN++ \cite{TextBPN++}, we employ a lightweight transformer network as our boundary refinement module (BRM). To simplify the model structure and reduce parameters, we adopt the strategy of sharing parameters and self-iterative refinement, as shown in Fig.~\ref{fig:framework}. In each refinement iteration, the BRM takes the previously predicted boundary as input and generates a new boundary, enabling dynamic refinement of text boundaries to adapt to various text shapes and scales.

As illustrated in Fig.~\ref{fig:framework}, the refined text boundaries become closer to the actual text boundaries as the number of iterations increases. Unlike TextBPN++\cite{TextBPN++}, which uses point matching loss, we optimize this module by minimizing the Smooth L1 distance between each refined text boundary ($T = {c_{_0},..., c_{_i},..., c{_{_K-1}}}$) and its corresponding target ($\tilde{T} = {\tilde{c}{_0},..., \tilde{c}{_i},..., \tilde{c}{_{_K-1}}}$). Specifically, the loss $ \mathcal{L}_{BR} $ of this module is formulated as
\begin{gather}
	\mathcal{L}_{BR} = -\frac{1}{K*N}\sum_{j=0}^{N-1} \sum_{i=0}^{K-1} Smooth_{L1}(c_{_{(j, i)}}, \tilde{c}_{_i})),\label{BR}
\end{gather}
where the $ N $ is the number of iterations, $ c_{_{(j, i)}} $ is $ j $-th refined prediction of point $ c_{_{i}} $.

\subsection{Dynamic Sampling}
\begin{figure}[tbp]
	\begin{center}
		\includegraphics[width=0.96\linewidth]{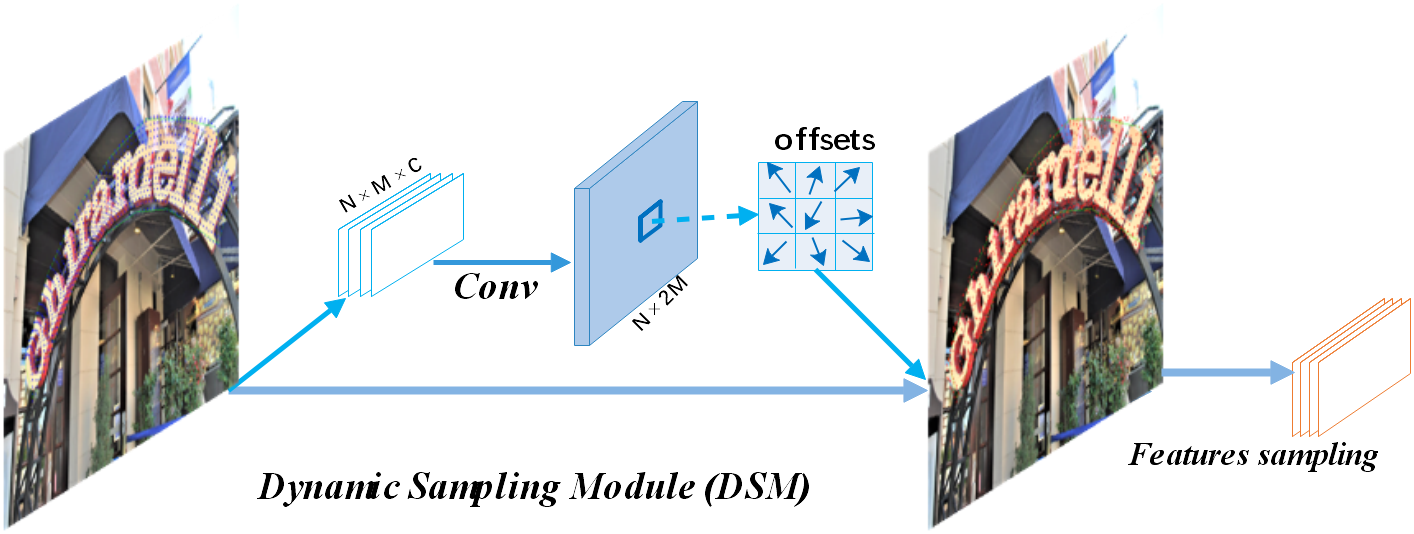}
		\caption{The schematic of the dynamic sampling module. The blue point is the sampling of TPS, and the red point is the sampling with dynamic adjustment.}	
		\label{fig:DTPS1}
	\end{center}%
	\vspace{-1.8em}
\end{figure}

Thin-Plate Spline (TPS) has been widely used as a grid sampling approach in arbitrary shape scene text spotting~\cite{TextPerception,Boundary_TextSpotter, TPSNet}. However, TPS is an inflexible sampling strategy that only generates fixed sampling grids manually, as illustrated by the blue grid points in Fig.~\ref{fig:DTPS1}. These fixed sampling grids result in the sampled CNN features for recognition sequence being highly dependent on the detected text boundary. As a consequence, it becomes challenging for the recognizer to decode the correct sequences when the detection is inaccurate or the sampling features are inadequate, as shown in Fig.~\ref{fig:intro1} (c1-c3). Therefore, the recognition performance highly relies on the text detection accuracy, leading to potential error propagation to recognition in these methods~\cite{TextPerception,Boundary_TextSpotter, TPSNet}.

To solve this problem, we propose a novel dynamic sampling module with thin-plate spline, named DSM,  which can dynamically sample the features in the detected text region for the recognition module. Our DSM can mitigate the incompatibility between text detection and recognition, especially when the detected text boundaries are imperfect for recognition. As shown in Fig.~\ref{fig:DTPS1}, the dynamic sampling module comprises two stacked $3\times 3$ convolutional layers with dilation [3,1] and one fully connected layer. Inspired by DCN\cite{DCN}, which learns offsets to produce deformable convolution kernels based on traditional convolution, we employ a lightweight convolution head to predict a set of position offsets for the basic grid points generated by the TPS algorithm. Specifically, we first use TPS to generate the fiducial grid points $G$ of size $w_o \times h_o$. Then, the lightweight convolution head produces a set of normalized position offsets for the fiducial grid points $G$, as 
\begin{gather}
	\Delta \hat{g}_{ij} = Tanh(C_{1 \times 1}(C_{3 \times 3}(C_{3 \times 3}(f_{ij}, d_3), d_1))),
\end{gather}
where $ C_{3 \times 3} $ denotes $ 3\times 3  $ convolutional layer, and $ C_{1 \times 1} $ denotes $ 3\times 3  $ fully connected layer realized by $ 1 \times 1$ convolution. $ f_{ij} $ is the input features of points $ {g}_{ij} $ extracted from recognition shared features ($ F_r $). $ d_3 $ and $ d_1 $ indicate that the dilation are set to 3 and 1, respectively. $ \Delta \hat{g}_{ij} $ is the normalized offset of the point $ g_{ij} $ in gird $  G $. To make the learning of offset $ \Delta \hat{g}_{ij} $  not affected by the size of the text instance, we normalize it by the sigmoid function. Therefore, the real offset can be calculated as 
\begin{gather}
	\Delta g_{ij}  = \Delta \hat{g}_{ij} \cdot o(w,h),\\
	G^{'} = G + \beta \cdot \Delta G, 
\end{gather}
where $ w $ and $ h $ are the width and height of the bounding rectangle of the text instance. $ \Delta G = \{\Delta g_{ij}|| i \in [0,w_o), j \in [0,h_o)\} $ is the set of $ \Delta g_{ij} $, and $ G^{'} $ is the updated gird. $ \beta $ is the scale coefficient and is set to 0.1, which can ensure a suitable offset space for the sampling points, as the detected text boundaries usually do not have significant errors.

\begin{figure}[tbp]
	\begin{center}
		\includegraphics[width=0.96\linewidth]{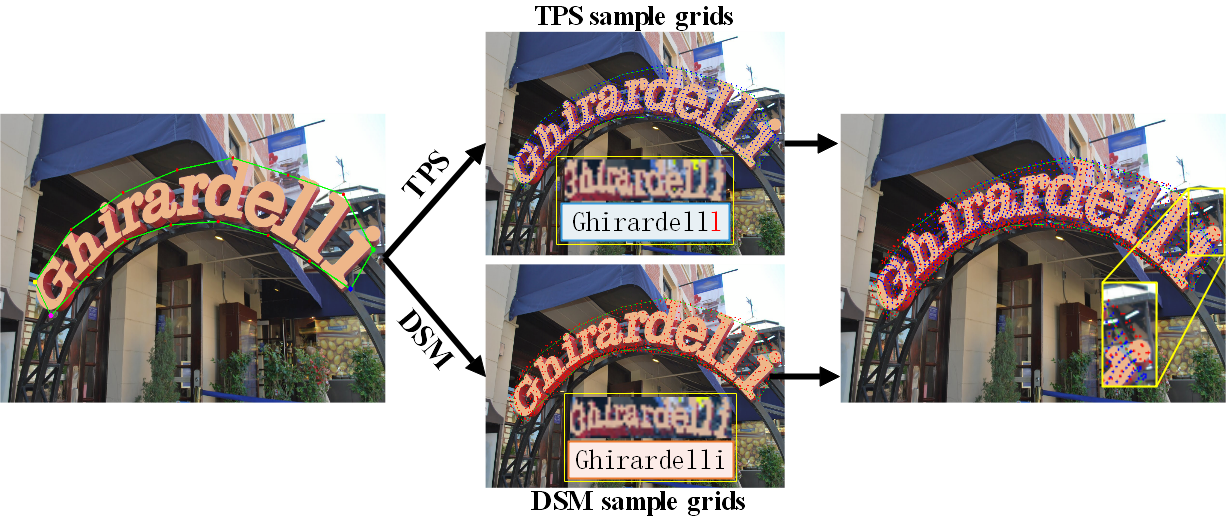}
		\caption{Visual comparison of TPS and DSM.}	
		\label{fig:DTPS2}
	\end{center}%
	\vspace{-1.8em}
\end{figure}

In training, DSM adaptively adjusts the position of fiducial grid points by using the gradient returned from the recognition module without extra supervision. During inference, DSM can dynamically sample appropriate features in the detected text region for the recognition module to decode the text sequence accurately. Our DSM can decode text content more accurately, even in challenging scenes where the detected text boundaries may have flaws or are imperfect for recognition, as shown in Fig.~\ref{fig:intro2} and Fig.~\ref{fig:DTPS2}. In Fig.~\ref{fig:DTPS2}, the detected text boundary does not cover the entire instance, making the recognition module decode the character `i' as `l' incorrectly. However, the sampling points of our DSM can exceed the detection boundary, enabling the recognition module to accurately decode the correct characters, even when the detected boundary does not completely cover the text instance.

\begin{table*}[tbp]
	\centering
	\caption{Ablations on test sets. ``	Rotation'' denotes training the model with random rotation augmentation. Without D-TPS means the TPS is used instead. Without REM means implicitly learning the reading-order only by the supervised regression learning of boundary control points, like~\cite{Boundary}.}
	\resizebox{0.92\linewidth}{!}{
		\begin{tabular}{cccc|ccccc|ccccc|ccccc}
			\toprule[1.5pt]
			\multirow{3}{*}{Method} &
			\multirow{3}{*}{Rotation} &
			\multirow{3}{*}{REM} &
			\multirow{3}{*}{DSM} &
			\multicolumn{5}{c|}{\textbf{Total-Text}} &
			\multicolumn{5}{c|}{\textbf{Rot.Total-Text}} &
			\multicolumn{5}{c}{\textbf{Inverse-Text}} \\
			\cmidrule(lr){5-9} \cmidrule(lr){10-14} \cmidrule(lr){15-19}
			&&& &\multicolumn{3}{c}{\textbf{Detection}}&\multicolumn{2}{c|}{\textbf{End-to-End}}&\multicolumn{3}{c}{\textbf{Detection}}&\multicolumn{2}{c|}{\textbf{End-to-End}}&\multicolumn{3}{c}{\textbf{Detection}}&\multicolumn{2}{c}{\textbf{End-to-End}}\\
			\cmidrule(lr){5-7}\cmidrule(lr){8-9} \cmidrule(lr){10-12}\cmidrule(lr){13-14} \cmidrule(lr){15-17}\cmidrule(lr){18-19}
			&&& &P&R&F& None & Full &P&R&F& None & Full &P&R&F& None & Full \\
			\midrule[1.1pt]
			\multirow{4}{*}{$Baseline$}& & &  &88.9 &82.6 &85.6 & 67.2 & 78.8 &80.8 &73.4 &76.9& 48.3 & 62.7 &81.9 &76.6 &79.2&52.6&66.6 \\
			
			&\checkmark & &  &89.9 &82.8 &86.2 & 67.4 & 79.2 &88.5 &81.8 &85.0& 51.8 & 66.2 &88.0 &81.8 &84.8&57.4&68.4 \\
			
			&\checkmark & &\checkmark  &90.8 &83.3 &86.9 & 69.6 & 80.7 &88.8 &83.2 &85.9& 53.4 & 68.0 &89.0 &82.7 &85.7&60.5&70.7 \\
			
			&\checkmark &\checkmark &  &91.2 &84.0 &87.5 & 69.2 & 80.3 &\textbf{88.9} &82.8 &86.2& 66.4 & 77.6 &88.6 &\textbf{83.8} &86.1&64.8&76.6 \\
		
			&\checkmark &\checkmark  &\checkmark  &\textbf{92.7} &\textbf{84.8} &\textbf{88.6} & \textbf{70.5} & \textbf{81.6} &89.4 &\textbf{84.6} &\textbf{86.9}& \textbf{68.8} & \textbf{80.2} &\textbf{90.3} &83.6 &\textbf{86.8}&\textbf{67.1}&\textbf{78.3}\\
			\bottomrule[1.5pt]
	\end{tabular}}
	\label{tab:main ablation}
	\vspace{-1.5em}
\end{table*}

\subsection{Optimization}

By dynamically sampling features by DSM, any recognition model can be applied for the recognition. For a fair comparison, we take the model in  \cite{ABCNet_v2, TPSNet} as our recognition module, which consists of 6 convolutional layers, one bidirectional LSTM layer, and an attention-based decoder. 

For end-to-end training the network, the objective
of the function is combined with the losses of modules mentioned above,
which is formulated as
\begin{gather}
	\mathcal{L} = \mathcal{L}_{IB} + {\lambda}_{re} *\mathcal{L}_{RE} + {\lambda}_{br} *\mathcal{L}_{BR} + {\lambda}_{rec} * \mathcal{L}_{REC},
\end{gather}
where $ \mathcal{L}_{IB} $ is the loss of the initial boundary module as in TextBPN++~\cite{TextBPN++};   $ \mathcal{L}_{RE} $ is the loss of reading-order estimation module as Eq.~\ref{RE}; $ \mathcal{L}_{BR} $ is the loss of boundary refinement module as Eq.~\ref{BR}; $ \mathcal{L}_{REC} $ is the Cross Entropy Loss for the recognition module as in \cite{ABCNet_v2, TPSNet}. In pre-training, the $ {\lambda}_{re}$ is set to $ {1}/{e^{(i-eps)/eps}} $, the $ {\lambda}_{br}$ is set to $ {0.1}/{e^{(i-eps)/eps}} $. $ eps $ denotes the maximum epoch of training, and $ i $ denote the $ i$-th epoch in training. In this way, our model can prioritise learning to find the interested text region at the beginning of training, ensuring that the training can converge normally. In fine-tuning, $ {\lambda}_{re}$ is set to 1.0, $ {\lambda}_{br}$ is set to $ 0.01$. $ {\lambda}_{rec} $ is empirically set to 0.2.

\section{Experiments} \label{Experiments}
\subsection{Datasets}

\noindent\textbf{SynthText 150k} is synthesized in~\cite{ABCNet} comes with 150k synthetic images containing mostly straight text and curved texts. It is different from SynthText 800k, which contains mostly straight texts in quadrilateral annotations.

\noindent\textbf{Total-Text} is a  word-level dataset including the horizontal, oriented and curved text, which contains 1255 training images and 300 test images.

\noindent\textbf{Rot.Total-Text}~\cite{DPText} is a test set derived
from Total-Text, which applies large rotation angles (0$ ^{\circ} $, 45$ ^{\circ} $, 135$ ^{\circ} $, 180$ ^{\circ} $, 225$ ^{\circ} $, 315$ ^{\circ} $) on images of the Total-Text test set to examine the model robustness, resulting in 1,800 test images.

\noindent\textbf{CTW-1500} is a line-level dataset containing horizontal, multi-oriented and curved text instances, including 1000 training images and 500
test images.

\noindent\textbf{ICDAR2015} is a word-level and multi-oriented text dataset, including 1000 training images and 500 test images.  This dataset includes many incidental scene text, such as blur or small text, which challenges text spotting.

\noindent\textbf{Inverse-Text}~\cite{DPText} is
established in~\cite{DPText}, only consists of 500
testing images.  It is a challenging arbitrarily-shaped scene text test set with about 40\% inverse-like scene texts, and some of these texts are even mirrored.



\subsection{Implementation Details}\label{trainstep}
We use ResNet-50 with DCN~\cite{DCN} as the backbone and pre-train our model for 100 epochs on a mixture of SynthText 150K, MLT-2017, Total-Text, and ICDAR2015 datasets. During pre-training, we apply the Adam optimizer with an initial learning rate of $ 0.001 $ and weight decay of $ 0.0001 $. The pre-trained model is then fine-tuned on the target dataset for 800 epochs with the initial learning rate set to $ 0.0001 $ and divided by ten at 400 epochs, using the Adam optimizer. We use 16 control points for text boundary and perform five iterations of BRM. The default number of sampling points is $ 16 \times 64 $, and data augmentation techniques such as random scaling, cropping, and distortion (e.g., random blur, brightness adjustment, and color change) are applied. To enhance the model's ability to recognize text in different reading-orders, we add random rotation with a wide-angle ($-180^{\circ}, 180^{\circ}$)) and denser sampling around [$ 0^{\circ}, \pm 30^{\circ}, \pm 60^{\circ}, \pm 90^{\circ}, \pm 120^{\circ}, \pm 150^{\circ}, \pm 180^{\circ} $].

During training, we randomly crop text regions without cutting any text and resize them to $ 640 \times 640 $. We pre-train our model using two RTX-3090 GPUs with an image batch size of 24 and fine-tune using a single RTX-3090 GPU with an image batch size of 12. During inference, we maintain the aspect ratio of test images and resize and pad them to the same size. Our implementation is based on PyTorch 1.7 and Python 3, and testing is performed on a single RTX-3090 GPU with a single thread. Recall, Precision, and F-measure are represented by R", P", and ``F", respectively.

\subsection{Ablation Studies}\label{exp_ablation}
In exploration experiments, we only pre-train the model 
with ten epochs to reduce training time costs. Other training details are described in Sec.~\ref{trainstep}. Some special details are described in the corresponding section. In default, random rotation data augmentation is used.

\begin{figure*}[htbp]
	\begin{center}
	\includegraphics[width=0.95\linewidth]{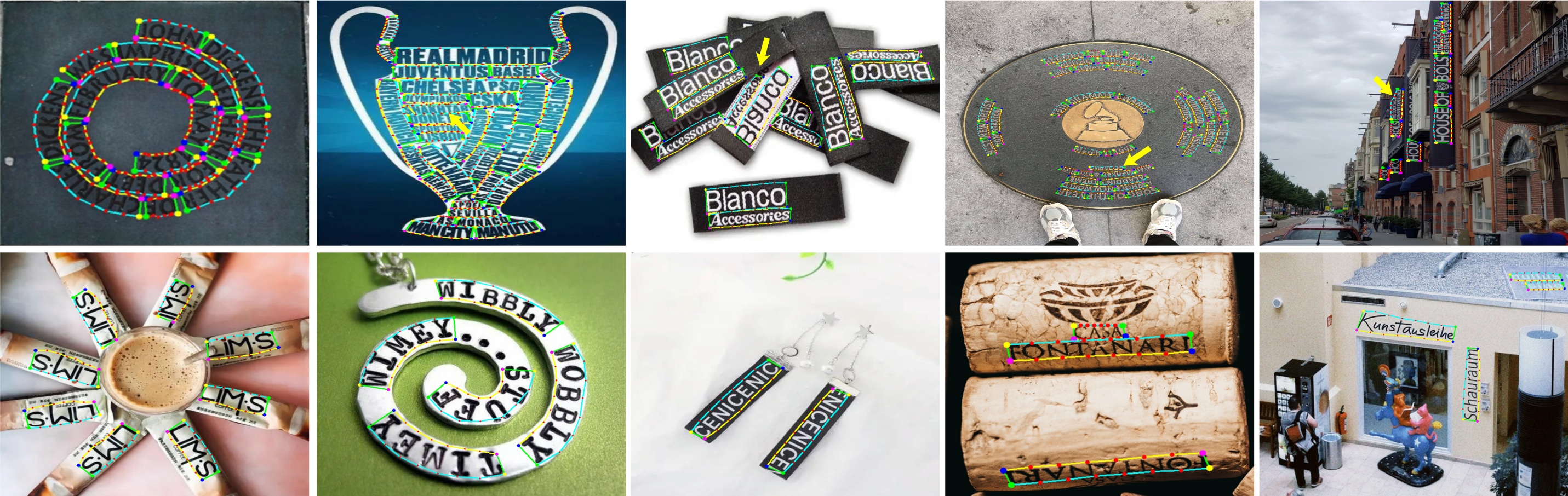}
	\caption{Qualitative results for inverse-like scene text detection. Our method can not only adapt to arbitrary shape text detection but also accurately mark the reading-order of each text instance. Some failure cases are marked with yellow arrows.}
	\label{fig:experiments1}
	\end{center}
\vspace{-2.0em}
\end{figure*}  

\begin{table}[t]
	\centering
	\caption{Results of spotters on Inverse-Text.} 
	\label{tab:ablation_sample}
	\resizebox{0.9\hsize}{!}{
		\begin{tabular}{lcccccc}
			\toprule[1.5pt]
			\multirow{2}{*}{Sampling Method} &
			\multirow{2}{*}{REM} &
			\multicolumn{3}{c}{Detection}&
			\multicolumn{2}{c}{End-to-End} \\
			\cmidrule(lr){3-5}\cmidrule(lr){6-7}
			&&P &R &F&None &Full \\
			\midrule[1.1pt]
			Masking ROI &&88.4 &80.4 &84.2&55.8&66.4 \\ 
			BezierAlign &&\textbf{89.3} &81.2&85.1&58.3&69.2\\
			TPS &&88.0 &81.8&84.8&57.4&68.4\\
			DSM &&89.0 &\textbf{82.7} &\textbf{85.7}&\textbf{60.5}&\textbf{70.7}\\
			
			\midrule[1.0pt]
			Masking ROI &\checkmark &89.0 &81.6 &85.1&62.4&74.2 \\ 
			BezierAlign &\checkmark&\textbf{90.5} &81.8 &85.9&65.0&77.2\\
			TPS &\checkmark&88.6 &\textbf{83.8}&86.1&64.8&76.6\\
			
			DSM &\checkmark&90.3 &83.6 &\textbf{86.8}&\textbf{67.1}&\textbf{78.3}\\
			\bottomrule[1.5pt]
	\end{tabular}}\\
\vspace{-1.0em}
\end{table}

\begin{figure}[tbp]
	\begin{center}
		\includegraphics[width=0.96\linewidth]{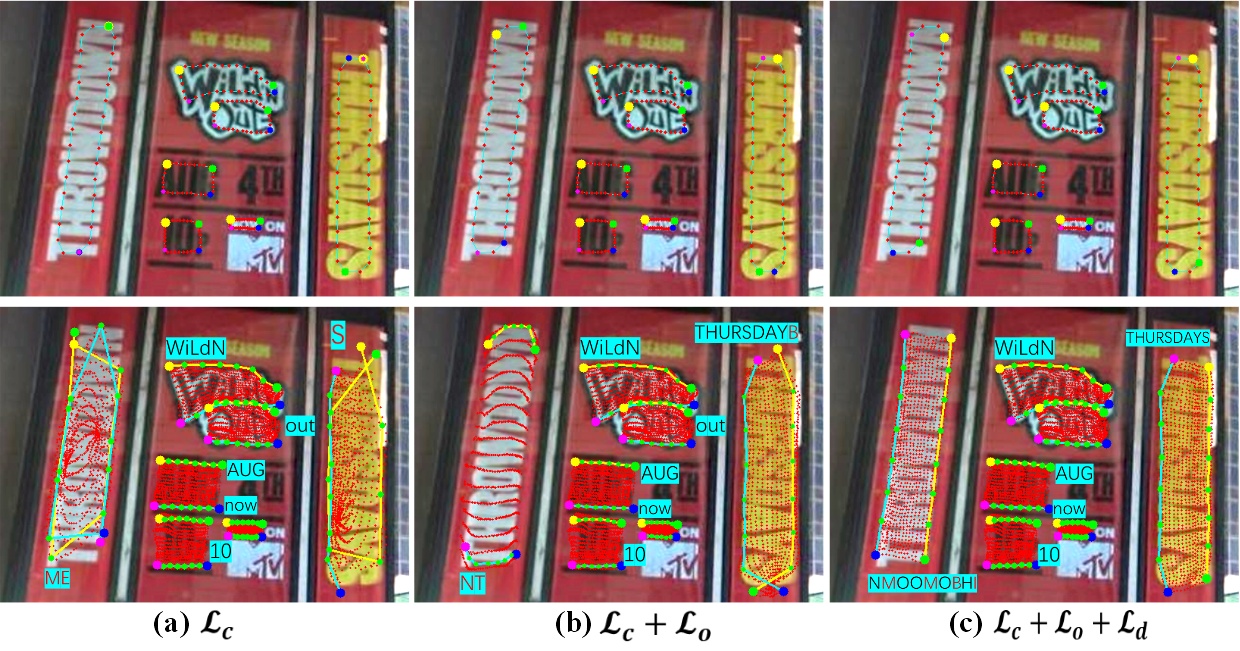}
		\caption{The intermediate visual comparison results for classification loss $ \mathcal{L}_{c} $, an orthogonality constraint $ \mathcal{L}_{o} $, and a distribution constraint $ \mathcal{L}_{d} $.}	
		\label{fig:loss_re}
	\end{center}%
	\vspace{-1.8em}
\end{figure}

\textbf{Reading-order estimation module (REM).} Tab.~\ref{tab:main ablation} shows that using the reading-order estimation module (REM) improves both detection and recognition performance, especially on Rot.Total-Text and Inverse-Text datasets, compared to the baseline model without REM. REM also brings significant improvement in text spotting performance, with 14.6\% and 11.4\% improvement on `None' and `Full', respectively, on Rot.Total-Text, and 7.4\% and 8.2\% improvement on `None' and `Full' respectively on Inverse-Text. However, the gain brought by REM is less significant on Total-Text, with only 1.8\% and 1.1\% improvement on `None' and `Full', respectively. The experimental results suggest that the impact of REM on text spotting performance depends on the ratio of inverse-like text in the testing data, which is about 40\% in Inverse-Text and relatively high in Rot.Total-Text due to small angle rotations that can turn some texts into inverse-like. The ratio of inverse-like instances is low in Total-Text, but REM is still important for accurately recognizing texts, as shown in Tab.~\ref{tab:ablation_sample}. Regardless of the sampling method, REM improves text recognition performance by about 6.5\% on `None' and 8\% on `Full' on Inverse-Text.

\begin{table}[htbp]
	\centering
	\caption{Results of spotters on Inverse-Text.} 
	\label{tab:ablation_reloss}
	\resizebox{0.9\hsize}{!}{
		\begin{tabular}{lcccccccc}
			\toprule[1.5pt]
			\multirow{2}{*}{$ \mathcal{L}_{RE} $} &
			\multirow{2}{*}{$ \mathcal{L}_{c} $}&\multirow{2}{*}{$ \mathcal{L}_{o} $} & \multirow{2}{*}{$ \mathcal{L}_{d} $}&
			\multicolumn{3}{c}{Detection}
			&\multicolumn{2}{c}{End-to-End} \\
			\cmidrule(lr){5-7}\cmidrule(lr){8-9}
			&&&&P&R&F&None &Full \\
			\midrule[1.1pt]
			+ classification&\checkmark & &&89.2 &82.1 &85.5&64.5&76.3 \\ 
			+ orthogonal&\checkmark &\checkmark&&89.7 &83.2 &86.3&66.4&77.4\\
			+ distribution&\checkmark &\checkmark &\checkmark&\textbf{90.3} &\textbf{83.6} &\textbf{86.8}&\textbf{67.1}&\textbf{78.3}\\
			\bottomrule[1.5pt]
	\end{tabular}}
\vspace{-1.2em}
\end{table}

\textbf{Joint reading-order estimation loss ($ \mathcal{L}_{RE} $).} 
In Tab.~\ref{tab:ablation_reloss}, we examine the impact of the proposed joint reading-order estimation loss ($ \mathcal{L}{RE} $) on Inverse-Text, which comprises a classification loss $ \mathcal{L}{c} $, an orthogonality constraint $ \mathcal{L}{o} $, and a distribution constraint $ \mathcal{L}{d} $. By conducting incremental experiments, we evaluate the effectiveness of each component in $ \mathcal{L}{RE} $. Our results reveal that when the orthogonality constraint $ \mathcal{L}{o} $ is applied, there is a performance improvement of 0.8\% in F-measure and 1.9\% in `None' for both detection and recognition tasks. Similarly, when we add the distribution constraint $ \mathcal{L}_{d} $, we observe a 0.5\% improvement in F-measure and 0.9\% in `Full' for both tasks. Combining $ \mathcal{L}{o} $ and $ \mathcal{L}{d} $ leads to significant performance improvement of 1.5\% in Recall, 1.3\% in F-measure, 2.6\% in `None', and 2.0\% in `Full'. We also present intermediate visual comparison results to verify the impact of orthogonality and distribution constraints on the prediction of reading-order, as depicted in Fig.~\ref{fig:loss_re} (a-c). Specifically, $ \mathcal{L}{o} $ ensures the independence of the four key points of reading-order and maintains their distinctness. In contrast, $ \mathcal{L}{d} $ constrains the orderly distribution of these points at the corners of the text boundary. The combination of three losses ($ \mathcal{L}{c} $, $ \mathcal{L}{o} $, and $ \mathcal{L}_{d} $) significantly improves the detection and recognition performance, as shown in Fig.~\ref{fig:loss_re} (c).

\begin{figure}[tbp]
	\begin{center}
    \includegraphics[width=0.96\linewidth]{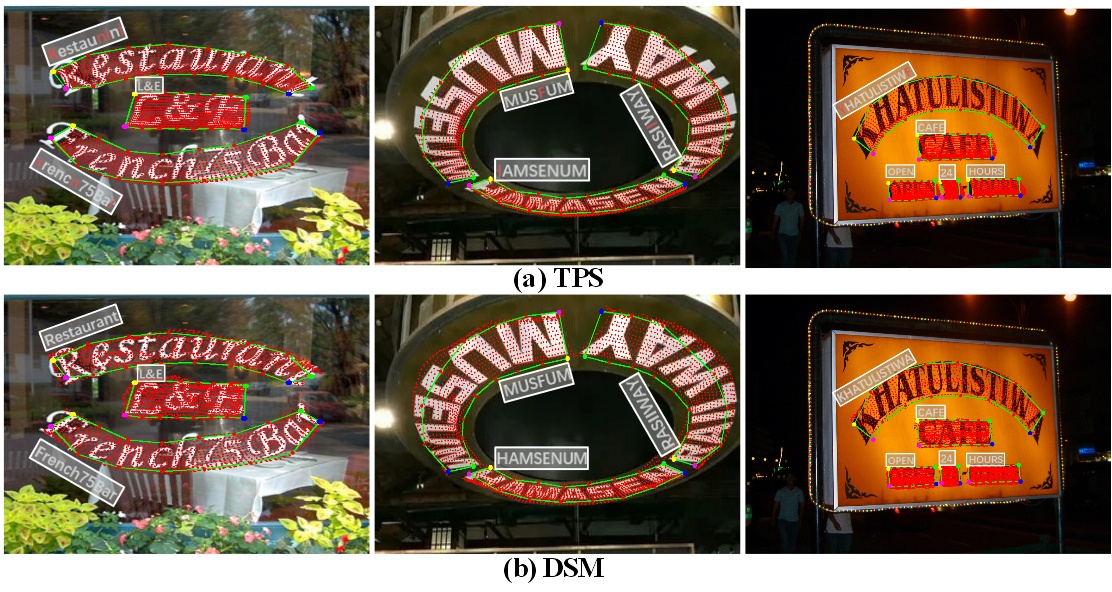}
		\caption{The intermediate visual comparison results for sampling methods; (a) TPS suffer from the precision of detected boundary, (b) DSM can adaptively adjust sampling points to alleviate the impact of detected boundaries.}	
		\label{fig:exp-dtps}
	\end{center}%
	\vspace{-1.5em}
\end{figure}

\textbf{Sampling methods.}
In Tab.\ref{tab:ablation_sample}, we can see that our DSM sampling method outperforms other methods such as Masking ROI, BezierAlign, and TPS. Specifically, with REM, our DSM improves performance by 4.7\% on `None' and 3.1\% on `Full' compared to Masking ROI, by 2.1\% on `None' and 1.1\% on `Full' compared to BezierAlign, and by 2.3\% on `None' and 1.7\% on `Full' compared to TPS. When the detection results are slightly poor without REM, the advantages of our DSM become even more significant. This demonstrates that our DSM performs well when text detection results are imperfect and can improve text detection. Moreover, our DSM also brings slight performance improvements in detection compared to TPS. Fig.~\ref{fig:exp-dtps} shows visual comparisons between TPS and DSM sampling points.

\begin{table}[tbp]
	\centering
	\caption{Analysis on initial boundary module (IBM) based on  detection results on Total-Text.} 
	\label{tab:ablation_ibm}
	\resizebox{0.9\hsize}{!}{
		\begin{tabular}{lccc}
			\toprule[1.5pt]
			\multirow{2}{*}{Method for initial Text Boundary} &
			\multicolumn{3}{c}{Detection Only} \\
			\cmidrule(lr){2-4}
			&Precision &Recall &F-measure \\
			\midrule[1.1pt]
			Text Kernel(PSENet)&89.59 &85.08 &87.28 \\ 
			Probability Map(DBNet) &\textbf{90.78} &84.81  &87.69\\
			Distance Field(TextBPN++) &90.51 &\textbf{85.50} &\textbf{87.93}\\
			\bottomrule[1.5pt]
	\end{tabular}}
\vspace{-1.5em}
\end{table}

\newif\ifnoFPS
\noFPStrue
\ifnoFPS

\begin{table*}[!t]
	\centering
		\caption{Scene text spotting results on Total-Text. 
			. $ *  $ denotes the method using ResNet50 with DCN as backbone. ``None'' represents lexicon-free, while ``Full'' indicates all the words in the test set are used.
			IC13 means ICDAR2013; IC15 means ICDAR2015; TT means Total-Text; MLT means MLT-2017.}
	\newcommand{\tabincell}[2]{\begin{tabular}{@{}#1@{}}#2\end{tabular}}
	\footnotesize
 \resizebox{0.95\linewidth}{!}{
	\begin{tabular}{ m{3.05cm}<{}
			|m{4.7cm}<{\centering}
			|m{2.6cm}<{\centering}
			|m{1.05cm}<{\centering}
			|m{0.39cm}<{\centering}
			|m{0.39cm}<{\centering}
			|m{0.39cm}<{\centering}
			|m{0.39cm}<{\centering}
			|m{0.39cm}<{\centering}
			|m{0.39cm}<{\centering}}
		
		\hline
		\multirow{2}*{Method}  & \multirow{2}*{Data} & \multirow{2}*{Backbone} & \multirow{2}*{Published} 
		& \multicolumn{3}{c|}{Detection} & \multicolumn{2}{c|}{End-to-end} & \multirow{2}*{FPS}\\
		\cline{5-9}
		&   &   & &P &R &F & None & Full & \\
		\hline
		Mask TextSpotter V1 \cite{MaskTextSpotter} & Syn800k, IC13, IC15, TT & ResNet-50-FPN &ECCV'18 &   69.0 &55.0 &61.3 & 52.9 & 71.8 & 4.8 \\
		\cline{1-4}
		CharNet \cite{CharNet} & Syn800k, IC15, MLT, TT & \tabincell{c}{ResNet-50-Hourglass57} &ICCV'19 &   87.3 &85.0 &86.1 &66.2 & - & 1.2 \\
		\cline{1-4}
		TextDragon \cite{TextDragon} & Syn800k, IC15, TT & \tabincell{c}{VGG16} &ICCV'19 & 85.6 &75.7 &80.3 & 48.8 & 74.8 & - \\
		\cline{1-4}
		TUTS \cite{TUTS} & \tabincell{c}{Syn200k, IC15, COCO-Text, TT,MLT, $ {\dagger} $} & ResNet-50-MSF &ICCV'19 &  83.3 &83.4 &83.3 & 67.8 & - & 4.8 \\
		\cline{1-4}
		Mask TextSpotter V3 \cite{Mask_TextSpotter_v3} & Syn800k, IC13, IC15, TT, AddF2k & ResNet-50-FPN   &ECCV'20 &-   &-   &- & 65.3 & 77.4 & 2.0 \\
		\cline{1-4}
		Text Perceptron~\cite{TextPerception}& Syn800k, IC13, IC15, TT\ & ResNet-50-FPN &AAAI'20 &88.8 &81.8 &85.2 & 69.7 & 78.3 &-\\
		\cline{1-4}
		ABCNet V1 \cite{ABCNet} & Syn150k, COCO-Text, TT, MLT & \tabincell{c}{ResNet-50-FPN} &CVPR'20 &-   &-   &- & 48.8 & 74.8 & - \\
		\cline{1-4}
		CRAFTS \cite{PGNet} & Syn800k,TT IC13, IC15& \tabincell{c}{ResNet-50-FPN} &ECCV'20 &89.5 &85.4 &87.4 & \textbf{78.7}  & - & - \\
		\cline{1-4}
		PGNet \cite{PGNet} & Syn150k, COCO-Text, TT, MLT & \tabincell{c}{ResNet-50-FPN} &AAAI'21 &85.5 & \textbf{86.8} &86.1  & 63.1  & - & 36 \\
		\cline{1-4}
		Boundary TextSpotter \cite{Boundary_TextSpotter} & Syn150k, COCO-Text, TT, MLT & \tabincell{c}{ResNet-50-FPN} &TIP'22 & 89.6 &81.2 &85.2 & 66.2  & 78.4 & 13 \\
		\cline{1-4}
		Li et al. \cite{TSNS} & Syn800k, IC13, IC15, TT, MLT, AddF2k & ResNet-101-FPN &TPAMI'22 &-   &-   &- & 57.8 & - & 1.4 \\
		\cline{1-4}
		PAN++ \cite{PAN++} & Syn150k, COCO-Text, TT, MLT & \tabincell{c}{ResNet-18-BFPN} &TPAMI'22 & 89.9 &81.0 &85.3 & 68.6 & 78.6 & 21 \\
		\cline{1-4}
		ABCNet V2 \cite{ABCNet_v2} & Syn150k, COCO-Text, TT, MLT LSVT& \tabincell{c}{ResNet-50-FPN} &TPAMI'22 &   
		90.2 &84.1 &87.0 &70.4 &78.1& 10 \\
		\cline{1-4}
		SPTS \cite{SPTS} & Syn150k, COCO-Text, TT, MLT & \tabincell{c}{ResNet-50-Transformer} &MM'22 &-   &-   &- & 74.2 & 82.4 & -\\
		\cline{1-4}
		TPSNet \cite{TPSNet} & Syn150k, TT, MLT, ArT & \tabincell{c}{ResNet-50-FPN$^{*}$}  &MM'22 &90.2 & \bf 86.8  &88.5 & \underline{76.1} & 82.3 & 9.3\\
		\cline{1-4}
		TESTR \cite{TESTR} & Syn150k, TT, MLT, IC15, IC13& \tabincell{c}{ResNet-50-Transformer} &CVPR'22 &  93.4 &81.4 &86.9 & 73.3 & \textbf{83.9} & 5.3\\
		\hline

		\hline
		\hline
		{\bf IAST}(Ours)  & \multirow{1}*{Syn150k, TT, MLT,  IC15} & \multirow{1}*{ResNet-50-FPN$^{*}$} &\multirow{1}*{-} &\textbf{94.7} &\underline{85.2} &\textbf{89.7} & 71.9 & \underline{83.5} & {7.8} \\
		\cline{1-1}
		\cline{5-10}
		\hline
\end{tabular}}
\label{tab:tt}
\vspace{-1.5em}
\end{table*}

\textbf{Initial boundary module (IBM).} We conducted experiments on the Total-Text dataset to assess the robustness of our model with different initial boundary generation strategies (such as Text Kernel in PSENet~\cite{CVPR19_PSENet}, Probability map in DB~\cite{DB}, Distance Field in TextBPN++\cite{TextBPN++}). To ensure a fair comparison, we trained the detection branch with the same settings (600 epochs with AdamW optimizer and initial learning rate of $0.0001$). Our results, as shown in Tab. \ref{tab:ablation_ibm}, indicate that different initial boundary generation strategies do not have a significant impact on detection performance (87.28\% F-measure for Text Kernel, 87.69\% F-measure for Probability map, 87.93\% F-measure for Distance Field). In our method, the initial boundaries are mainly used to locate the text instance roughly, and the subsequent boundary refinement module refines these coarse boundaries to accurate text boundaries. The detected text boundary of any text detection method can be used as our initial boundary, meaning that any text detection method (such as TextField, DB, PSENet, \etc) can serve as our IBM to quickly build an end-to-end text spotting method, even if its detection results are not as satisfactory.


\begin{table}[t]
	\centering
	\caption{Results of detection on Total-Text. and ``Iter 0'' indicates our experiment just using initial text boundary as detection.} 
	\label{tab:ablation_iter}
	\resizebox{0.85\linewidth}{!}{
		\begin{tabular}{lcccc}
			\toprule[1.5pt]
			\multirow{2}{*}{Iteration} &
			\multicolumn{3}{c}{Detection Only} \\
			\cmidrule(lr){2-5}
			&Precision &Recall &F-measure &FPS\\
			\midrule[1.1pt]
			Iter. 0 &\textbf{91.21} &74.09 &81.76& 10.87\\ 
			Iter. 1 &91.15 &82.44 &86.58& 10.64\\
			Iter. 2 &91.04 &84.62 &87.71&10.42\\
			Iter. 3 &90.51 &85.50 &\textbf{87.93}&10.21\\
		    Iter. 4 &90.45 &\textbf{85.53} &87.92&10.02\\
		    Iter. 5 &90.40 &85.50 &87.88&9.80\\
			\bottomrule[1.5pt]
	\end{tabular}}
\vspace{-1.8em}
\end{table}

\textbf{Refining iteration number.}
Our BRM model allows flexibility in choosing the number of iterations during testing, thanks to its shared parameters and self-iterative design. In Tab.~\ref{tab:ablation_iter}, we demonstrate that as the number of iterations increases, the detection performance gradually improves and eventually stabilizes while the inference speed decreases. Notably, even a single iteration leads to a significant improvement in detection performance. With three iterations, the performance stabilizes around 87.9\% on F-measure. The approximate time cost for each iteration is about 2ms, much less than the time cost of other parts (about 92 ms). Balancing efficiency and performance, we set the default number of iterations to 3 during testing.


\subsection{Comparisons with State-of-the-art Methods}\label{exp_Comparisons}
We evaluate our method on four publicly available benchmarks: Total-Text, CTW-1500, ICDAR2015, and Inverse-Text. Quantitative results against other state-of-the-arts are presented in Tab. \ref{tab:tt}, \ref{tab:icdar2015}, \ref{tab:ctw1500}, and \ref{tab:inversetext}. For inverse-like scene text detection and spotting, we show qualitative visual results in Fig.~\ref{fig:experiments1} and \ref{fig:experiments-E2E}, respectively.

\begin{table}[!t]
	\centering
	\caption{\small End-to-end text spotting results on CTW-1500. ``None'' represents lexicon-free, while ``Full'' indicates all the words in the test set are used.}\label{tab:ctw1500}
	\vspace{-0.3em}
	\resizebox{0.85\linewidth}{!}{%
		\begin{tabular}{l ccc cc}
			\toprule[1.5pt]
			\multirow{2}{*}{Method} & \multicolumn{3}{c}{Detection} & \multicolumn{2}{c}{End-to-End} \\ \cmidrule(lr){2-4} \cmidrule(lr){5-6} 
			& P & R & F & None & Full  \\ 
			\midrule
			TextDragon\cite{TextDragon} & 84.5 & 82.8 & 83.6 & 39.7 & 72.4 \\ 
			Text Perceptron\cite{TextPerception} & 87.5 & 81.9 & 84.6 & 57.0 & $-$  \\
			ABCNet\cite{ABCNet} & $-$ & $-$ & $-$ & 45.2 & 74.1  \\ 
			ABCNet v2\cite{ABCNet_v2} & 85.6 & 83.8 & 84.7 &  57.5 & 77.2 \\
			MANGO\cite{MANGO} & $-$ & $-$ & $-$ & 58.9 & 78.7  \\
			TESTR-Bezier~\cite{TESTR} & \underline{89.7} & 83.1 & 86.3 & 53.3 & 79.9  \\ 
			TESTR-Polygon~\cite{TESTR} & \textbf{92.0} & 82.6 & \textbf{87.1} & 56.0 & 81.5 \\
			SPTS-Bezier~\cite{SPTS} & $-$ & $-$ & $-$ &52.6 &73.9 \\
			SPTS-Point~\cite{SPTS} & $-$ & $-$ & $-$ &\textbf{63.6} &\textbf{83.8} \\
			TPSNet~\cite{TPSNet} &87.7 &\textbf{85.1}  &86.4 &59.7 &79.2 \\
			ABINet++~\cite{ABINet++} & $-$ & $-$ & $-$ &60.2 &80.3 \\
			\hline
			{\bf IAST}(Ours) & 89.2& \underline{84.8} & \underline{86.9} & \underline{62.4} & \underline{82.9}  \\ 
			\bottomrule[1.5pt]
		\end{tabular}%
	}
	\vspace{-1.8em}
\end{table}

\textbf{Total-Text}. In testing, we scale the input image sides into (640, 1024) while maintaining the aspect ratio. As shown in Tab. \ref{tab:tt}, our method outperforms all previous methods in detection, surpassing the best-reported result by 4.5\% on Precision and 1.2\% on F-measure. For the end-to-end case, our method achieves competitive results (83.5\% on `Full') compared to methods based on "ResNet-50-Transformer" like TESTR~\cite{TESTR} (83.9\% on `Full'). Our method outperforms all previous methods based on "ResNet-50-FPN" and surpasses the best-reported result~\cite{TPSNet} by 1.2\% on `Full'.

\begin{table}[tbp]
	\caption{\small Results on ICDAR2015 dataset. ``S'', ``W'', ``G'', ``N'' represent recognition with ``Strong'', ``Weak'', ``Generic'' or ``None'' lexicon respectively.}\label{tab:icdar2015}
	\centering
	\setlength{\tabcolsep}{3pt}
	\resizebox{0.85\linewidth}{!}{%
		\begin{tabular}{l ccc ccc c}
			\toprule[1.5pt]
			\multirow{2}{*}{Method} & \multicolumn{3}{c}{Detection} & \multicolumn{4}{c}{End-to-End} \\
			\cmidrule(lr){2-4} \cmidrule(lr){5-8}
			& P & R & F & S & W & G & N \\
			\midrule
			TextNet\cite{TextNet} & 89.4 & 85.4 & 87.4 & 78.7 & 74.9 & 60.5 & $-$ \\ 
			FOTS\cite{FOTS} & 91.0 & 85.2 & 88.0 & 81.1 & 75.9 & 60.8 & $-$ \\
			CharNet R-50\cite{CharNet} & 91.2 & \underline{88.3} & \underline{89.7} & 80.1 & 74.5 & 62.2 & 60.7 \\
			Boundary\cite{Boundary} & 89.8 & 87.5 & 88.6 & 79.7 & 75.2 & 64.1 & $-$ \\
			TUTS\cite{TUTS} & 89.4 & 85.8 & 87.5 & 83.4 & \underline{79.9} & 68.0 & $-$ \\
			Text Perceptron\cite{TextPerception} & \underline{92.3} & 82.5 & 87.1 & 80.5 & 76.6 & 65.1 & $-$ \\
			Mask TextSpotter v3\cite{Mask_TextSpotter_v3} & $-$ & $-$ & $-$ & 83.3 & 78.1 & \textbf{74.2} & $-$ \\
			Boundary TextSpotter\cite{Boundary_TextSpotter} &88.7 &84.6 &86.6 &82.5 &77.4 &71.7& $-$\\
			ABCNet v2\cite{ABCNet_v2} &  90.4 & 86.0 & 88.1 & 82.7 & 78.5 & 73.0 & $-$  \\
			MANGO\cite{MANGO} & $-$ & $-$ & $-$ & 81.8 & 78.9 & 67.3 & $-$ \\
			PGNet\cite{PGNet} & 91.8 & 84.8 & 88.2 & 83.3 & 78.3 & 63.5 & $-$ \\
			PAN++~\cite{PAN++} &91.4 &83.9 &87.5 &82.7 &78.2 &69.2 & $-$\\
			TESTR~\cite{TESTR} & 90.3 & \textbf{89.7} & \textbf{90.0} & \textbf{85.2} & 79.4 & 73.6 & \textbf{65.3}  \\
			SPTS~\cite{SPTS}&$-$ & $-$ & $-$ &77.5 &70.2 &65.8&$-$\\
			\midrule
			{\bf IAST}(Ours) & \textbf{92.5} & 86.6 & 89.5& \underline{84.4}& \textbf{80.0} & \underline{73.8} & \underline{64.7} \\
			\bottomrule[1.5pt]
		\end{tabular}}
\vspace{-1.5em}
\end{table}

\begin{figure*}[htbp]
    \begin{center}
   \includegraphics[width=0.95\linewidth]{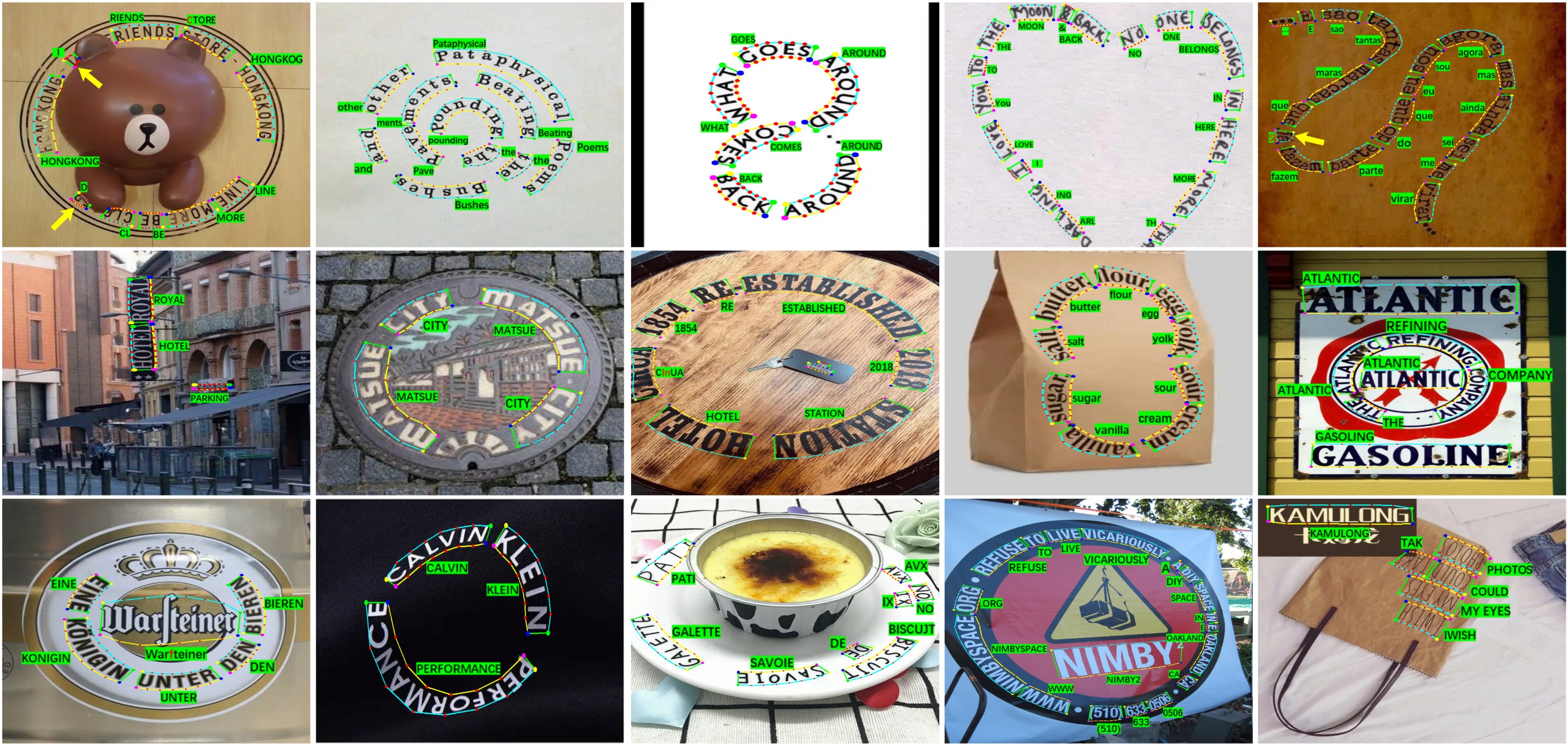}
	\caption{Qualitative results for inverse-like scene text spotting on Inverse-Text. Our method can accurately read various complex layout and inverse-like scene texts. Some failure cases are also marked with yellow arrows.}
	\label{fig:experiments-E2E}
	\end{center}
\vspace{-1.6em}
\end{figure*} 

\textbf{CTW-1500}. For CTW-1500, we set the number of control points on the text boundary to 32, and the number of sampling points is set to $16\times128$ due to the line-level curved text dataset. The input image sides are scaled into (640, 1024) while maintaining the aspect ratio. As shown in Tab. \ref{tab:ctw1500}, our method outperforms all previous methods in both detection and end-to-end results, except for CNN-Transformer-based methods like TESTR~\cite{TESTR} and SPTS~\cite{SPTS}. Specifically, our method outperforms all previous CNN-based methods like ABINet++\cite{ABINet++}, TPSNet~\cite{TPSNet}, ABCNet v2~\cite{ABCNet_v2}, and surpasses the best-reported result~\cite{ABINet++} by 2.2\% on `None' and 2.6\% on `Full' in terms of F-measure. In detection, our method only slightly lags behind the best-reported CNN-Transformer-based method~\cite{TESTR} by 0.2\% on F-measure. Our method ranks second in most metrics, with only a slight gap to the first. Due to CTW-1500 is a line-level dataset, there are some overly long sentences inside. Our method based on CNN is limited by the receptive field, which may result in lower recognition accuracy than these transformer based method. Another problem is that CTW-1500 has some unreasonable or missing annotations, as mentioned in~\cite{TextPMs, DRRG}, which also can bringing some performance losses.

\textbf{ICDAR2015}. We conducted evaluations on the ICDAR2015 benchmark, which includes many perspective texts, and the results are listed in Tab.\ref{tab:icdar2015}. During testing, we scaled the short side of the input image to 960 while maintaining its aspect ratio. Our method achieved comparable performance with text spotting methods like CharNet\cite{CharNet} and TESTR in the detection stage. In the Word Spotting tasks, our method also delivered a competitive performance with text spotting methods like Mask TextSpotter v3~\cite{Mask_TextSpotter_v3} and TESTR, and it achieved remarkable performance (80.0\%) on weak lexicon cases. Furthermore, our method significantly outperformed previous TPS-based methods like TUTS and `Boundary TextSpotter', demonstrating the effectiveness of our method.

\newif\ifnoFPS
\noFPStrue
\ifnoFPS

\begin{table}[t]
	\centering
	\caption{Results of spotters on Inverse-Text. ``DPText'' indicates the experimental results are gained from DPText~\cite{DPText}, ``repro''  indicates the experiment results are reproduced  by DPText based its official released code. $ * $ indicates the results are reported in DPText by testing the  officially  released model of previous state-of-the-art spotters on  Inverse-Text.}
	\label{tab:inversetext}
	\resizebox{\hsize}{!}{
		\begin{tabular}{lcccc cc}
			\toprule[1.5pt]
			\multirow{2}{*}{Method} &
			\multirow{2}{*}{Rotation} &
			\multirow{2}{*}{P} &
			\multirow{2}{*}{R} &
			\multirow{2}{*}{F} &
			\multicolumn{2}{c}{End-to-End} \\
			\cmidrule(lr){6-7}
			& & & & &None &Full \\
			\midrule[1.1pt]
			ABCNet v2$ * $ \cite{ABCNet_v2} &\checkmark&82.0 &70.2 &75.6 &34.5 &47.4 \\
			TESTR$ * $ \cite{TESTR} &\checkmark&83.1 &67.4 &74.4 &34.2 &41.6\\
			SwinTextSpotter$ * $ \cite{SwinTextSpotter} &\checkmark&94.5 &85.8 &89.9 &55.4 &67.9 \\
			ABCNet-v2 (DPText repro.) &\checkmark &83.4 &73.2 &78.0 &57.2 &69.5 \\
			ABCNet-v2 w/ Pos.Label (DPText repro.) &\checkmark &90.7 &83.9 &87.2 &62.2 &\underline{76.7} \\
			TESTR (DPText repro.) &\checkmark &89.4 &84.4 &86.8 &62.1 &74.7\\
			TESTR w/ Pos.Label (DPText repro.) &\checkmark &88.8 &85.7 &87.2 &61.9 &74.1\\
			TESTR w/ Pos.Label (DPText detector) &\checkmark &90.7 &84.2 &87.3 &63.1 &75.4\\
			SwinTextSpotter (DPText repro.) &\checkmark &\textbf{94.5} &84.7 &\underline{89.3} &62.9 &74.7 \\
            SPTS~\cite{SPTS} &\checkmark &- &-&- &38.3&46.2 \\
            DeepSolo (Res-50,\#2)~\cite{DeepSolo} &\checkmark &- &-&- &64.6 &71.2 \\
            DeepSolo (ViTAEv2-S,\#3)~\cite{DeepSolo} &\checkmark &- &-&- &\underline{68.8} &75.8 \\
		\midrule
		{\bf IAST}(Ours) &\checkmark &\underline{92.5} &\textbf{86.6} &\textbf{89.5} &\textbf{68.8} &\textbf{80.6} \\ 
		   \bottomrule[1.5pt]
	\end{tabular}}
\vspace{-1.5em}
\end{table}

\begin{figure*}[htbp]
\begin{center}
\includegraphics[width=0.95\linewidth]{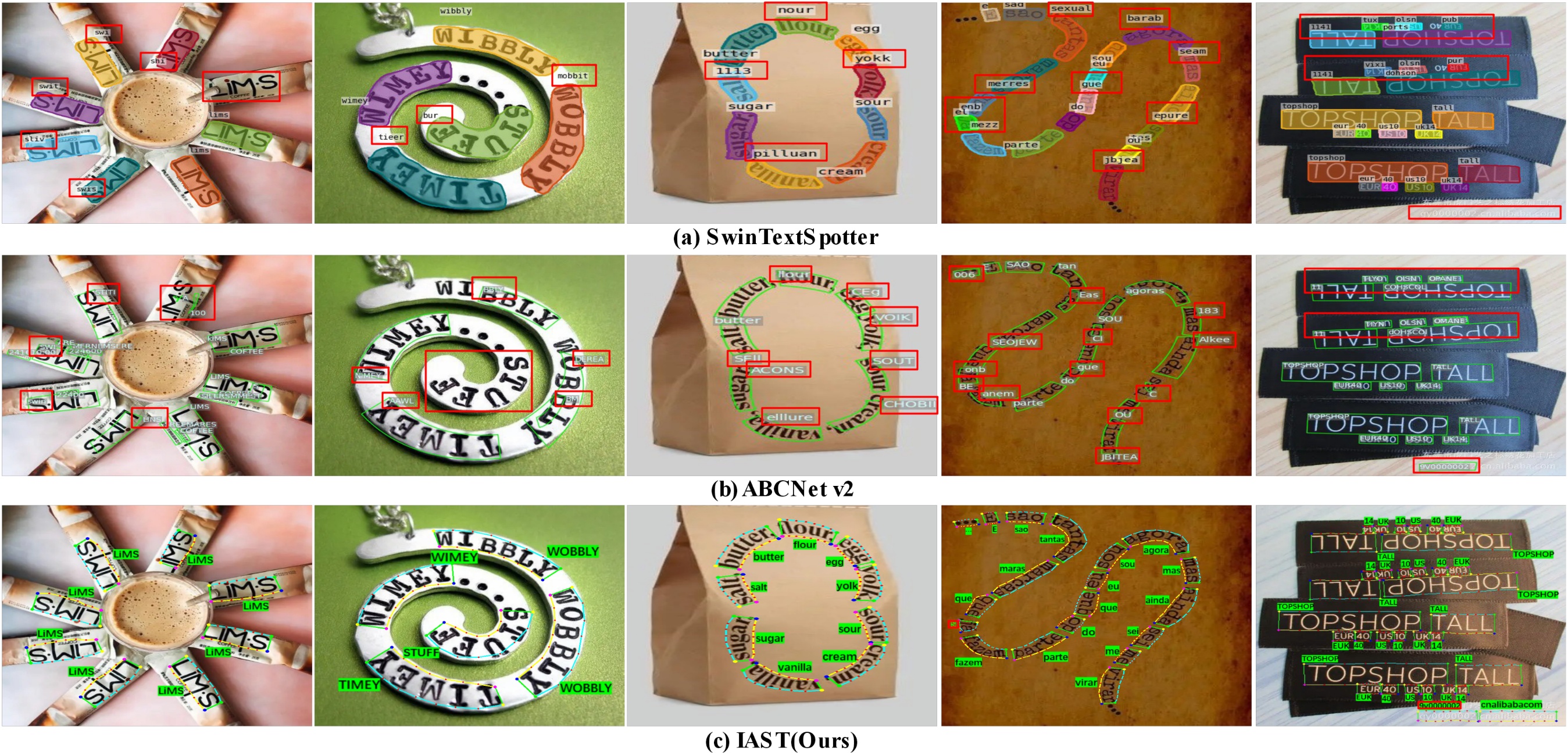}
\caption{Some visual comparison results with ABCNet v2~\cite{ABCNet_v2} and SwinTextSpotter~\cite{SwinTextSpotter} on Inverse-Text dataset, where The results of ABCNet v2 and SwinTextSpotter are reproduced by their official open-source code and model. Some failure cases of each method are highlighted in red boxes.}
\label{fig:experiments-com}
\end{center}
\vspace{-1.8em}
\end{figure*} 

\textbf{Inverse-Text}. Although reading-order is important for scene text spotting, the ratios of inverse-like scene texts are relatively low in several datasets, such as 2.8\% in Total-Text, 5.2\% in CTW-1500, and 0.0\% in ICDAR2015. Therefore, in these datasets, our reading-order estimation module (REM) does not fully demonstrate our DSM module achieves its advantages and more positive gains. To demonstrate the effectiveness of our REM, we conducted experiments on Inverse-Text, an arbitrary-shape scene text test set with approximately 40\% inverse-like instances. As Inverse-Text does not have a training set, we fine-tuned our model on Total-Text with random
rotation for 800 epochs and evaluated it on Inverse-Text, scaling the input image to (640, 1280) while maintaining its aspect ratio. Our method outperformed all previous methods and achieved the best performance on multiple evaluation metrics (\eg, Recall, F-measure, None, Full). In the text spotting task, our method surpassed the best result by 5.7\% on `None' and by 3.9\% on `Full', significantly outperforming CNN-Transformer based methods such as TESTR and SwinTextSpotter. Furthermore, our IAST accurately predicts the reading-order of scene text with very complex layouts and recognizes inverse-like scene text via the predicted reading-order, as shown in Fig.~\ref{fig:experiments1} and Fig.~\ref{fig:experiments-E2E}. Our results demonstrate the effectiveness of our methods in reading inverse-like scene text and highlight the importance of reading-order for scene text spotting, particularly in complex layout scenes.

\textbf{Visual Comparison.} We utilize some previous state-of-the-art spotters to get qualitative results on Inverse-text for giving a more
intermediate visual comparison, as shown in Fig.~\ref{fig:experiments-com}. We select ABCNet-v2~\cite{ABCNet_v2} representing the CNN-based spotting methods which predict control points. We select SwinTextSpotter~\cite{SwinTextSpotter} representing the
transformer-based spotting methods which predict text segmentation. Officially released model weights trained on Total-Text are adopted for producing visual results. We can find that the spotters (\eg, ABCNet-v2 and SwinTextSpotter) failed to correctly recognize inverse-like texts because of the unconventional reading-order, as shown in Fig.~\ref{fig:experiments-com} (a) and (b). Benefiting from the ability to aware reading-order, our model can  accurately recognize these inverse-like texts, as shown in Fig.~\ref{fig:experiments-com} (c). These examples in Fig.~\ref{fig:experiments-com} are enough to demonstrate the improvement of our method when reading the complex layout and inverse-like texts.

\begin{figure}[tbp]
	\begin{center}
		\includegraphics[width=0.90\linewidth]{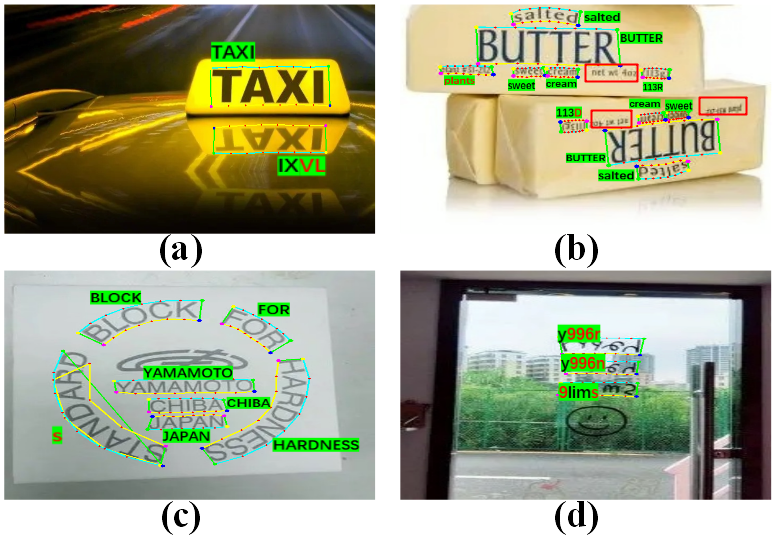}
		\caption{Failures cases on hard inverse-like texts, such as mirror text, overlap of key points, and small blurry text with red boxes.}	
		\label{fig:exp-weak}
	\end{center}%
	\vspace{-1.8em}
\end{figure}

\subsection{Weakness}
The proposed method shows strong performance in spotting both normal and inverse-like scene text in most cases, although it may struggle with some special cases, such as mirror text, small blurry text, and occlusion text, as shown in Fig.~\ref{fig:exp-weak}. Some failure cases are also marked with yellow arrows in Fig.~\ref{fig:experiments1} and Fig.~\ref{fig:experiments-E2E}. As shown in Fig.~\ref{fig:exp-weak} (a) and (d), mirror texts are especially challenging as their control points need to be pressed counterclockwise. Additionally, some small blurry texts are detected, but their recognition accuracy cannot be verified, as shown in Fig.~\ref{fig:exp-weak} (b), as they were annotated as `Don't Care'. In rare cases, overlapping key points in the reading-order may result in unsatisfactory predictions, as shown in Fig.~\ref{fig:exp-weak} (c). Fortunately, these errors can be corrected through additional processing. It's worth noting that these special cases are challenging for all text spotting methods and require further research.

\section{Conclusion} \label{Conclusion}
Inverse-like text is a common problem in scene text recognition, but it has not been attracted enough attention and effectively solved. To address this problem, we propose a unified end-to-end trainable framework called IAST, which accurately reads both normal and inverse-like scene text through reading-order estimation and dynamic sampling. Our REM module can learn and extract reading-order information from the initial text boundary, while the DSM can dynamically sample appropriate features for recognition in the detected text region. Extensive experiments demonstrate the effectiveness of our methods for reading inverse-like scene text and highlight the importance of reading-order information for scene text spotting, especially in complex layouts.

\bibliographystyle{IEEEtran}
\bibliography{IEEEbib}

 \end{document}